%% file: arxiv.tex
\documentclass[final]{cvpr}

\input{header}

\graphicspath{{figures/}}

\usepackage[pagebackref=true,breaklinks=true,colorlinks,bookmarks=false]{hyperref}

\def\cvprPaperID{5745} 
\def\confYear{CVPR 2021}

\begin{document}

\title{DeepMetaHandles: Learning Deformation Meta-Handles of 3D Meshes with Biharmonic Coordinates}

\author{Minghua Liu\textsuperscript{1} $\quad$
Minhyuk Sung\textsuperscript{2} $\quad$
Radomir Mech\textsuperscript{3} $\quad$
Hao Su\textsuperscript{1} \\
\textsuperscript{1}University of California San Diego $\quad$ \textsuperscript{2}KAIST $\quad$ \textsuperscript{3}Adobe Research\\
}

\makeatletter
\let\@oldmaketitle\@maketitle
\renewcommand{\@maketitle}{\@oldmaketitle
    \vspace{-2\baselineskip}
    \includegraphics[width=\linewidth]{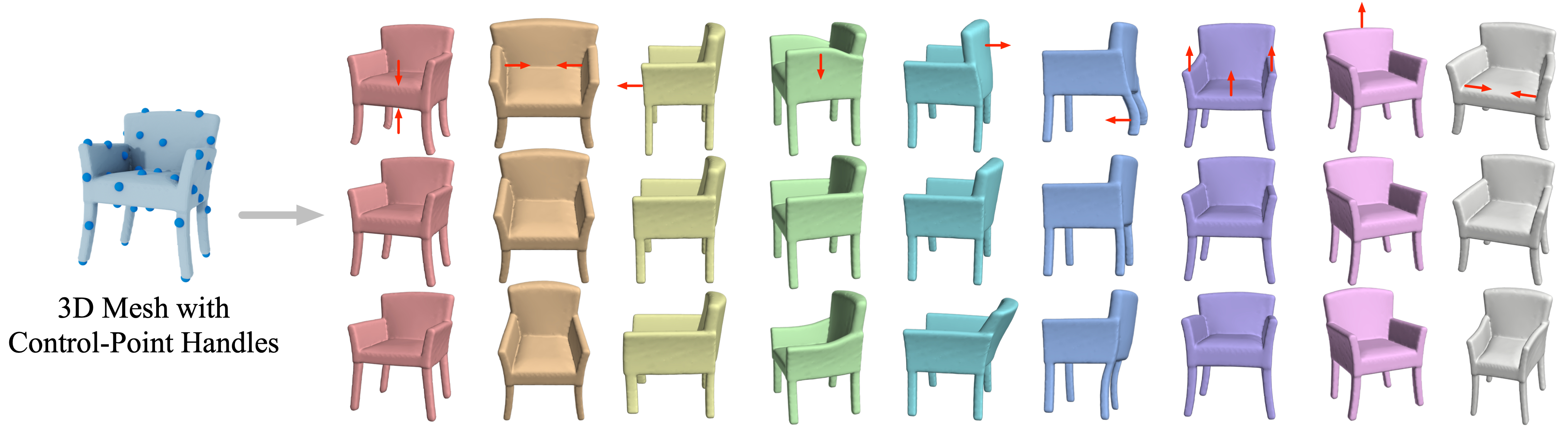}
    \vspace{-6mm}
    \captionof{figure}{Learned meta-handles for a single chair. Each column indicates a meta-handle and shows three deformations along the direction of that meta-handle, with red arrows highlighting the deformed region. Our method learns intuitive and disentangled meta-handles in an unsupervised fashion, which factorize all the plausible deformations for the shape.}
    \label{fig:teaser}
\bigskip}
\makeatother

\maketitle


\begin{abstract}
\input{sections/abstract}

\end{abstract}

\input{sections/introduction}
\input{sections/related_work}
\input{sections/method}

\input{sections/results}

\input{sections/conclusion}

\paragraph{Acknowledgments} This work is supported in part by gifts from Adobe, Kwai, Qualcomm, and Vivo.

{\small
\bibliographystyle{ieee_fullname}
\bibliography{egbib}
}
   
\clearpage
 
\renewcommand{\thesection}{S}
\setcounter{table}{0}
\renewcommand{\thetable}{S\arabic{table}}
\setcounter{figure}{0}
\renewcommand{\thefigure}{S\arabic{figure}}
 
\newif\ifpaper 
\papertrue 

\section{Supplementary Material}
\input{sections/supplementary}

\end{document}

%% file: header.tex
\usepackage{amsfonts,amsmath,amssymb}
\usepackage{bbm}
\usepackage{booktabs}
\usepackage{caption}
\usepackage{color}
\usepackage{comment}
\usepackage{dsfont}
\usepackage{enumitem}
\usepackage{epsfig}
\usepackage{graphicx}
\usepackage{makecell}
\usepackage{multirow}
\usepackage{tabularx}
\usepackage{wrapfig}
\usepackage{xspace}
\usepackage{times}
\usepackage[font=small]{caption}
\usepackage{enumitem}

\let\originalparagraph\paragraph
\renewcommand{\paragraph}[2][.]{\originalparagraph{#2#1}}

\newcolumntype{L}{>{\raggedright\arraybackslash}X}
\newcolumntype{C}{>{\centering\arraybackslash}X}

\makeatletter
\DeclareRobustCommand\onedot{\futurelet\@let@token\@onedot}
\def\@onedot{\ifx\@let@token.\else.\null\fi\xspace}

\def\etal{\emph{et al}\onedot}
\makeatother

\makeatletter
\newcommand\footnoteref[1]{\protected@xdef\@thefnmark{\ref{#1}}\@footnotemark}
\makeatother

%% file: sections/abstract.tex
\vspace{-5pt}

We propose \textbf{DeepMetaHandles}, a 3D conditional generative model based on mesh deformation. Given a collection of 3D meshes of a category and their deformation handles (control points), our method learns a set of meta-handles for each shape, which are represented as combinations of the given handles. The disentangled meta-handles factorize all the plausible deformations of the shape, while each of them corresponds to an intuitive deformation. A new deformation can then be generated by sampling the coefficients of the meta-handles in a specific range. We employ biharmonic coordinates as the deformation function, which can smoothly propagate the control points' translations to the entire mesh. To avoid learning zero deformation as meta-handles, we incorporate a target-fitting module which deforms the input mesh to match a random target. To enhance deformations' plausibility, we employ a soft-rasterizer-based discriminator that projects the meshes to a 2D space. Our experiments demonstrate the superiority of the generated deformations as well as the interpretability and consistency of the learned meta-handles. The code is available at \url{https://github.com/Colin97/DeepMetaHandles}.

%% file: sections/introduction.tex
\vspace{-15pt}
\section{Introduction}

\vspace{-3pt}

3D Meshes can store sharp edges and smooth surfaces compactly. However, Learning to generate 3D meshes is much more challenging than 2D images due to the irregularity of mesh data structures and the difficulty in designing loss functions to measure geometrical and topological properties. For such reasons, to create new meshes, instead of generating a mesh from scratch, recent work assumes that the connectivity structure of geometries is known so that the creation space is restricted to changing the geometry without altering the structure. For example, \cite{Tan:2018a,Tan:2018b,zhou2020unsupervised} create new shapes by \emph{deformations} of one template mesh. They, however, limit the scope of the shape generation to possible variants of the template mesh.  We thus propose a 3D \emph{conditional} generative model that can take any existing mesh as input and produce its plausible variants.  Our approach integrates a \emph{target-driven} fitting component and a conditional generative model. At test time, it allows both deforming the input shape to fit the given target shape and exploring plausible variants of the input shape without a target. 

Our main design goals are two-fold: improving the \emph{plausibility} of the output shapes and enhancing the \emph{interpretability} of the learned latent spaces. To achieve the goals, the key is to choose a suitable parameterization of deformations. One option is to follow the recent target-driven deformation network~\cite{3DN,CycleConsistency,NeuralCages,Sung:2020}, which parameterizes the deformation as new positions of all the mesh vertices. However, such a large degree of freedom often results in the loss of fine-grained geometric details and tends to cause undesirable distortions. Instead of following the above works, we leverage a classical idea in computational geometry, named \emph{deformation handles}, to parameterize smooth deformations with a low degree of freedom. Specifically, we propose to take a small set of \emph{control points} as deformation handles and utilize a deformation function defined on the control points and their \emph{biharmonic coordinates}~\cite{Wang:2015}.

Not all the translations of the control points lead to plausible deformations. Based on the control-point handles, we aim to learn a low-dimensional deformation subspace for each shape, and we expect the structure of this subspace to exhibit \emph{interpretability}. In contrast to typical generative models, where shape variations are embedded into a latent space implicitly, our method explicitly factorizes all the plausible deformations of a shape with a small number of interpretable deformation functions. Specifically, for each axis of our input-dependent latent space, we assign a deformation function defined with the given set of control points and offset vectors on them so that each axis corresponds to an intuitive deformation direction. Since each axis is explicitly linked to multiple control-point handles, we thus call them \emph{meta-handles}. We enforce the network to learn \emph{disentangled} meta-handles, in the sense that a meta-handle should not only leverage the correlations of the control-point handles, but also correspond to a group of parts that tend to deform altogether according to the dataset. We hope that the disentangled meta-handles allow us to deform each part group independently in downstream applications. 

Beyond choosing the parameterization of deformations, we have to overcome the challenge of examining the plausibility. In the popular adversarial learning framework, a straightforward approach would be converting the output mesh to voxels or point clouds and exploiting voxel or point cloud based discriminators. The conversions, however, may discard some important geometric details. In our method, we instead project the shapes into a 2D space with a differentiable \emph{soft rasterizer}~\cite{liu2019soft} and employ a 2D discriminator. We found that this architecture can be trained more robustly, and it captures local details of plausible shapes.

Our deformation-based conditional generative model, named \textbf{DeepMetaHandles}, takes random pairs of source and target shapes as input during training. For the source shape, the control points are sampled from its mesh vertices by farthest point sampling, and the biharmonic cooridnates~\cite{Wang:2015} for control-point handles are pre-computed. Our network consists of two main modules: MetaHandleNet and DeformNet. The MetaHandleNet first predicts a set of meta-handles for the source shape, where each meta-handle is represented as a combination of control-point offsets. A deformation range is also predicted for each meta-handle, describing the scope of plausible deformations along that direction. The learned meta-handles, together with the corresponding ranges, define a deformation subspace for the source shape. Then, DeformNet predicts coefficients multiplied to the meta-handles, within the predicted ranges, so that the source shape deformed with the coefficients can match the target shape. To ensure the plausibility of variations within the learned subspace, we then randomly sample coefficients within the predicted ranges and apply both geometric and adversarial regularizations to the corresponding deformations.

Fig.~\ref{fig:teaser} shows examples of the learned meta-handles, which interestingly resemble natural deformations of \emph{semantic} parts, such as lifting the armrests or bending the back of a chair. Our experiments also show that the learned meta-handles are consistent across various shapes and well disentangle the shape variation space. Finally, we compare our approach with other target-driven deformation techniques~\cite{huang2017learning,3DN,CycleConsistency,NeuralCages} and demonstrate that our method produces superior fitting results.

\originalparagraph{Key contributions:}
\vspace{-1\baselineskip}
\begin{itemize}[leftmargin=0.25cm]
    \setlength\itemsep{-0.5em}
    \vspace{-0.5\baselineskip}
    \item
    We propose DeepMetaHandles, a 3D \emph{conditional} generative model based on mesh deformation.
    \item
    We employ a few control points as deformation handles. Together with their biharmonic coordinates, we can produce smooth but flexible enough deformations.
    \item
    We propose to factorize the deformation space with a small number of disentangled meta-handles, each of which provides an intuitive deformation by leveraging the correlations between the control points.
    \item
    We improve the plausibility of the deformations by exploiting a differentiable renderer and a 2D discriminator.
\end{itemize}
\vspace{-1\baselineskip}

%% file: sections/related_work.tex
\section{Related Work}
\vspace{-3pt}

\paragraph{Learning 3D Shape Deformations}
3D shape deformation is a classic subject in computational geometry that has been studied extensively for decades. The problem is typically formulated as an optimization problem minimizing the fitting error from the source to target shape and also some regularization errors (e.g., local rigidity). Recent work, however, has demonstrated how neural networks can be leveraged in the shape deformation not only for improving the fitting accuracy but also for multiple other purposes such as: to fit the source shape to a partial target shape~\cite{Hanocka:2018} or 2D images~\cite{Jack:2018,Kurenkov:2018,3DN}, to find point-wise correspondences through deformation~\cite{groueix20183d,CycleConsistency}, to predict customized deformation handles for each input shape~\cite{NeuralCages}, to cluster shapes given a collection~\cite{Mehr:2019}, to learn semantic deformations~\cite{Yumer:2016}, and to transfer deformations~\cite{Yang:2018,Sung:2020}. While our approach can also perform target-driven deformation, our main goal is different: to learn a deformation-based conditional generative model. We also remark that our method does not utilize any semantic supervision such as part segmentation, as done by some recent works~\cite{Sung:2020,Yang:2020}.

\vspace{-1em}
\paragraph{3D Shape Generative Models}

In light of the success in the 2D image case, deep generative models have also been widely investigated for 3D data. Wu~\etal~\cite{Wu:2016} was the first proposing a 3D GAN with voxel representation, and Achlioptas~\etal~\cite{Achlioptas:2018} and their subsequent work~\cite{valsesia:2019,Shu:2019} also proposed point-cloud-based GANs. However, these approaches are not able to produce fine-grained geometric details due to the limit of the resolution.While mesh is a preferable representation, generating meshes is very challenging in particular when preventing the generation of non-manifold faces or disconnected components~\cite{Nash:2020}. Hence, previous work, such as the one of Tan~\etal~\cite{Tan:2018a,Tan:2018b}, considers generating novel shapes by deforming a given template mesh, limiting the scope of the generation to the possible variations of the template shape. We propose to overcome this limitation with our \emph{conditional} generative model, which takes any 3D mesh as input to deform. Generative models for 3D shapes have also been investigated to learn possible variations of compositional structures with or without semantic annotations~\cite{Gao:2019,Schor:2019,Dubrovina:2019,Wu:2020,Yang:2020,Li:2017, StructureNet}. In this work, we focus on learning geometric variations of the given shape while preserving its topological structure. 

%% file: sections/method.tex
\section{Method}
\vspace{-3pt}

In this section, we will first briefly review the control-point-based deformation and the biharmonic coordinates~\cite{Wang:2015} technique we use, and introduce how the meta-handles are defined with the control-point handles (Section~\ref{sec:biharmonic}). We will then present how we learn the meta-handles in an unsupervised fashion and our neural network architectures  (Section~\ref{sec:learning_meta_handles}). Lastly, we will introduce our loss functions that guide the emergence of plausible deformations and intuitive factorizations (Section~\ref{sec:loss_functions}).

\subsection{Biharmonic Coordinates and Meta-Handles}
\label{sec:biharmonic}
\vspace{-3pt}

Mesh deformation through directly moving individual vertex is cumbersome and may easily lead to unwanted distortions. We thus leverage deformation handles to parameterize the deformations with a low degree of freedom. The key in the handle-based deformation is to define a proper deformation function that features several desired properties. For instance, no change of handles should result in no deformation; each handle should produce local and smooth deformation; the deformation function should be expressed in closed form. Numerous previous work has introduced different handle-based deformation functions. Many of them are based on solving the \emph{biharmonic} equation defined over the mesh with boundary constraints (given from handles). The resulting deformation functions of these approaches satisfy many desired properties~\cite{Jacobson:2011,Jacobson:2012}. Also, closed-form expressions with respect to the handles can be easily calculated after a pre-computation. (Please refer to Jacobson~\etal~\cite{Jacobson:2014} for more details.)

In our method, we employ a subset of mesh vertices as the deformation handles (\emph{control points}) and restrict the transformations of the handles to pure translations. Given the mesh vertices $\mathbf{V}\in\mathbb{R}^{n \times 3}$ ($n$ vertices) and a set of $c$ control points $\mathbf{C}\in\mathbb{R}^{c \times 3}$, the \emph{linear} map $\mathbf{W}\in\mathbb{R}^{n \times c}$ between them ($\mathbf{V} = \mathbf{W}\mathbf{C}$) is often called `generalized barycentric coordinates'~\cite{Meyer:2002,Ju:2005,Joshi:2007,Manson:2010}. Wang~\etal~\cite{Wang:2015} proposed one way to define $\mathbf{W}$ based on the biharmonic functions, which is thus dubbed \emph{biharmonic coordinates}, and we utilize it as our deformation function. Without requiring that control points form a \emph{cage} enclosing the input shape, our deformation handles are flexible and intuitive.

Specifically, we sample $c$ control points from the mesh vertices by farthest point sampling (FPS) over the geodesic distances. The biharmonic coordinates $\mathbf{W}$ are also precomputed. However, the deformation function $f: \mathbb{R}^{c \times 3} \rightarrow \mathbb{R}^{n \times 3}$ defined over the given control points $\mathbf{C}$, $f(\mathbf{C}) = \mathbf{W}\mathbf{C}$, has $3c$ degrees of freedom. It may overparameterize the plausible shape variation space, which means there may be lots of implausible deformations, if we randomly translate the control points (see Fig.~\ref{fig:control-point}).  Also, there
 \begingroup
\setlength{\columnsep}{3mm}%
\begin{wrapfigure}{r}{0.5\linewidth}\vspace{-3mm}
   \includegraphics[width=\linewidth]{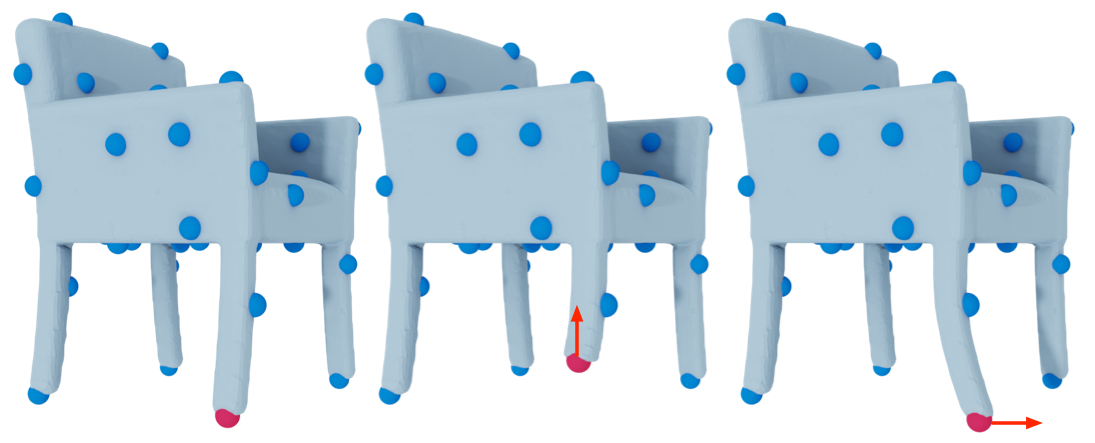}
\vspace{-7mm}
   \caption{Two deformations resulted from moving the red control point along the arrow directions.}
   \label{fig:control-point}
\vspace{-4mm}
\end{wrapfigure}
  may exist strong correlations across the deformations from moving individual control points. For a specific shape (e.g., a chair), all the plausible variants may reside in a lower-dimensional subspace and can be factorized into several meaningful deformation directions (e.g., scaling all chair legs and bending the chair back).

\endgroup

\begin{figure*}[t]
\begin{center}
   \includegraphics[width=\linewidth]{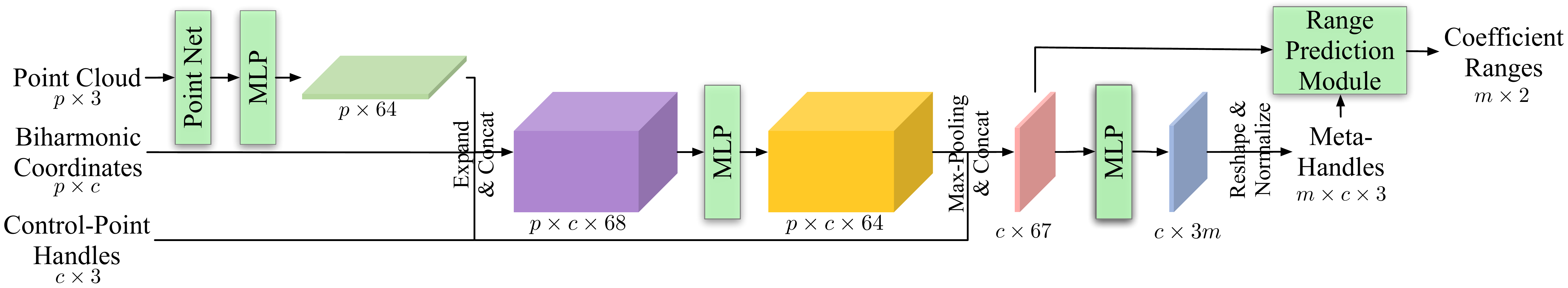}
\end{center}
\vspace{-\baselineskip}
\caption{Architecture of MetaHandleNet: it incorporates the information from the shape (point cloud), control-point handles, and biharmonic coordinates by building a 3D tensor, and predicts a set of meta-handles with the corresponding coefficient ranges for the shape.}
\label{fig:meta-handle-network}
\end{figure*}

\begin{figure}[t]
\begin{center}
   \includegraphics[width=\linewidth]{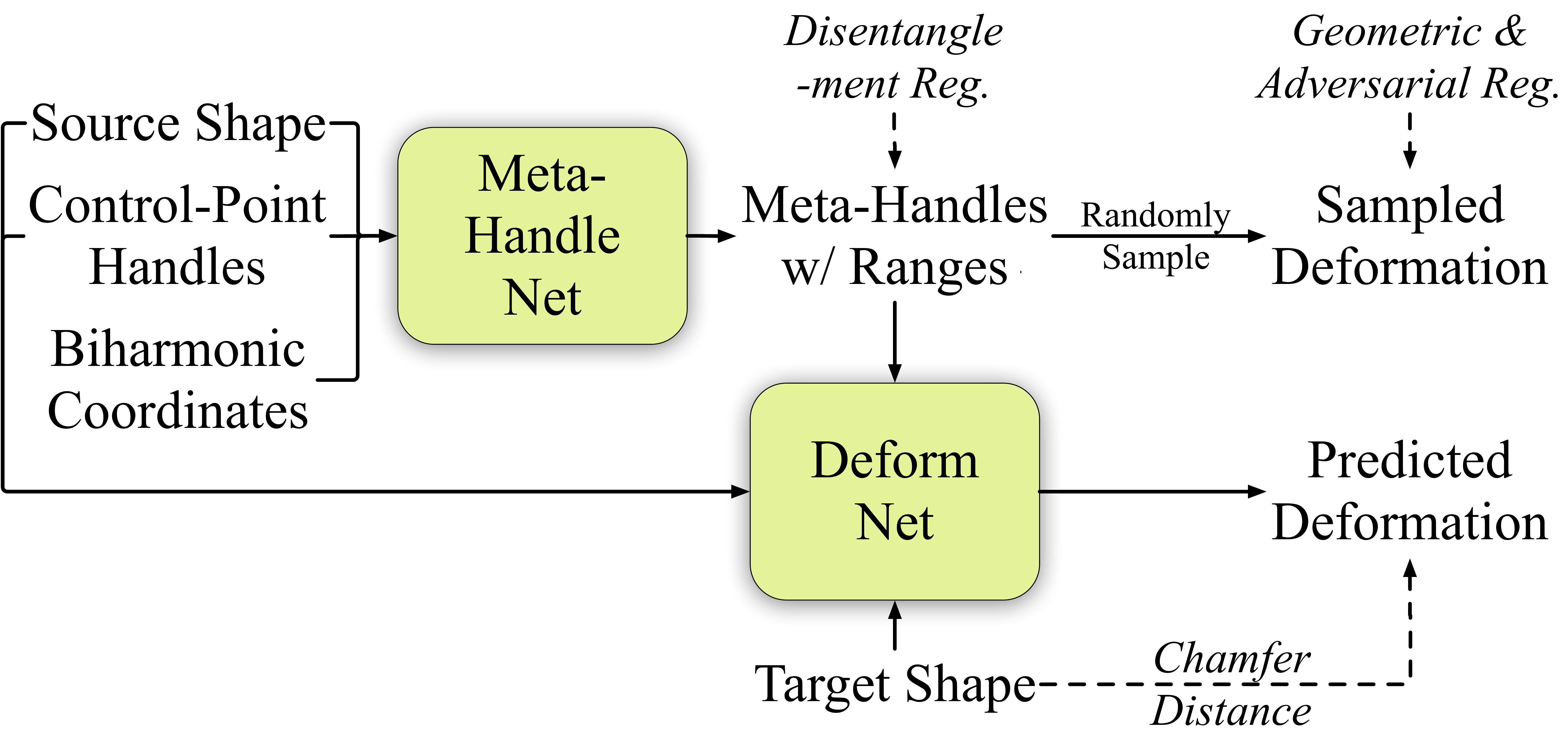}
\end{center}
\vspace{-\baselineskip}
\caption{Overview of our method. We learn the meta-handles in an unsupervised fashion.}
\label{fig:overview}
\end{figure}

To this end, we propose to find a smaller number of meta-handles to factorize the subspace covering all the plausible deformations. Specifically, each meta-handle $\mathbf{M}_i \in \mathbb{R}^{c \times 3}$ is represented as \emph{offsets} over the $c$ control points:
\vspace{-1.4mm}
\begin{equation}
     \mathbf{M}_i = [\vec{t}_{i1}, \cdots, \vec{t}_{ic}]^T,
\vspace{-1.4mm}
\end{equation}
where $\vec{t}_{ij} \in \mathbb{R}^3$ indicates the offset of the $j$-th control point for the $i$-th meta-handle. In contrast to a single control point that mainly affects a local region of the mesh, each meta-handle is expected to provide a more intuitive deformation direction, which may even correspond to some semantic meanings (See Figs.~\ref{fig:teaser} and~\ref{fig:multiple-shape-meta-handle}).

We now use the linear combination of the meta-handles to represent a deformation. Specifically, a new deformation function $g: \mathbb{R}^{m} \rightarrow \mathbb{R}^{n \times 3}$ is defined with respect to the meta-handles $\{ \mathbf{M}_i \}_{i=1 \cdots m}$ and their linear combination coefficients $\mathbf{a} = [a_1, \cdots, a_m]$:
\vspace{-2.2mm}
\begin{equation}
    g \left( \mathbf{a}; \{ \mathbf{M}_i \}_{i=1 \cdots m} \right) = \mathbf{W} (\mathbf{C}_0 + \sum_{i=1}^{m}a_i\mathbf{M}_i),
    \label{equ:func_g}
\vspace{-1.7mm}
\end{equation}
where $\mathbf{C}_0\in\mathbb{R}^{c \times 3}$ denotes the rest positions of the given control points. In the context of the conditional generative model, it can be interpreted as that each shape has a $m$-dimensional input-dependent latent space, where each axis corresponds to a meta-handle describing a specific deformation function in 3D space. A latent code $\mathbf{a}$ can thus be directly decoded to a deformation of the input mesh as a linear combination of the meta-handles.

Along with the meta-handles, our method also predicts \emph{ranges} $\{ [L_i, R_i] \}_{i=1 \cdots m}$ of the coefficients associated with each meta-handle. The ranges describe the scope of plausible deformations along the direction of each meta-handle. Any set of coefficients within the \emph{ranges} $\{ [L_i, R_i] \}_{i=1 \cdots m}$ is thus expected to produce a plausible deformation.

We utilize a small number of meta-handles to learn a low-dimensional compact deformation space. The degrees of freedom of the deformation function $g$ is typically much smaller than that of the deformation function $f$, i.e., $m \ll 3c$. As a result, the meta-handles are required to not only leverage the correlations of the control-point handles, but also discover the underlying properties of the shape structure (e.g., chair legs are symmetric and should thus be deformed together). 

\subsection{Network Architecture}
\label{sec:learning_meta_handles}
\vspace{-3pt}

We propose to learn the meta-handles in an unsupervised fashion without taking semantic annotations or correspondences across the shapes as input or supervision. As shown in Fig.~\ref{fig:overview}, our method mainly includes three networks: MetaHandleNet, DeformNet, and a discriminator network (discussed in Section~\ref{sec:loss_functions}). Taking a pair of randomly sampled shapes within the same category as input, the method predicts a deformation space for the source shape, and finds a deformation within the space to match the target shape. Specifically, MetaHandleNet takes a source shape, its control points, and the precomputed biharmonic coordinates as input and predicts a set of meta-handles as well as the corresponding coefficient ranges. DeformNet then predicts coefficients of the meta-handles so that the resulting deformation of the source shape matches the target shape.

To ease encoding, in MetaHandleNet, we convert the input source mesh to a point cloud (denoted as $\mathbf{P}\in\mathbb{R}^{p \times 3}$) by uniformly sampling $p$ points over the mesh surface. The precomputed biharmonic coordinates are also interpolated from the mesh vertices to the point cloud (i.e., $\mathbf{W}\in\mathbb{R}^{p \times c}$)  according to the barycentric coordinates. Fig.~\ref{fig:meta-handle-network} illustrates the architecture of MetaHandleNet. It first encodes the point cloud with PointNet~\cite{PointNet} and obtains 64-dimensional features per point, which is denoted as $\mathbf{D}\in\mathbb{R}^{p \times 64}$. Then, the point features $\mathbf{D}$, the biharmonic coordinates $\mathbf{W}$, and the rest positions of the control points $\mathbf{C_0}\in\mathbb{R}^{c \times 3}$ are consolidated in a 3D tensor (a purple volume in Fig.~\ref{fig:meta-handle-network}). Specifically, the 3D tensor has a size of $p \times c \times 68$, and the first $p \times c \times 64$ is packed with the point features $\mathbf{D}$ (repeating for the control points), the next $p \times c \times 1$ 
is filled with the biharmonic coordinates $\mathbf{W}$, and the last $p \times c \times 3$ is filled with the rest positions of the control points $\mathbf{C_0}$ (repeating for the point cloud). Hence, in this tensor, each pair of a point in $\mathbf{P}$ and a control point has a 68-dimensional feature, which is processed with an MLP. We then aggregate the features across the points through a symmetric function (i.e., max-pooling) to produce 64-dimensional features per control point. The control-point feature is combined again with the rest position of the control point and is then converted to a $3m$-dimensional vector through another MLP, which becomes the offsets for the $m$ meta-handles. We then normalize each metal-handle to unit length to facilitate training. The predicted meta-handles and the 67-dimensional control-point features are then fed into a range prediction module, which outputs a coefficient range $[L_i, R_i]$ for each meta-handle. Please refer to the supplementary materials for the details of the module.

As for the DeformNet, it takes the source shape, target shape, the predicted meta-handles with the coefficient ranges, and the control-point features (extracted from MetaHandleNet) as input, and predicts a coefficient vector $\mathbf{a} \in \mathbb{R}^m$ within the predicted ranges $\Pi_{i=1}^{m}[Li,Ri]$. The predicted coefficient vector and the meta-handles are then fed into the deformation function $g$ (Equation~\ref{equ:func_g}) to decode the deformation for the source shape, which is expected to match the target shape. Similar to MetaHandleNet, DeformNet also builds a 3D tensor to incorporate all the information and utilize shared-weight MLPs and max-pooling to process and aggregate the features. Please refer to the supplementary materials for the details.
\begin{figure}[t]
\begin{center}
   \includegraphics[width=\linewidth]{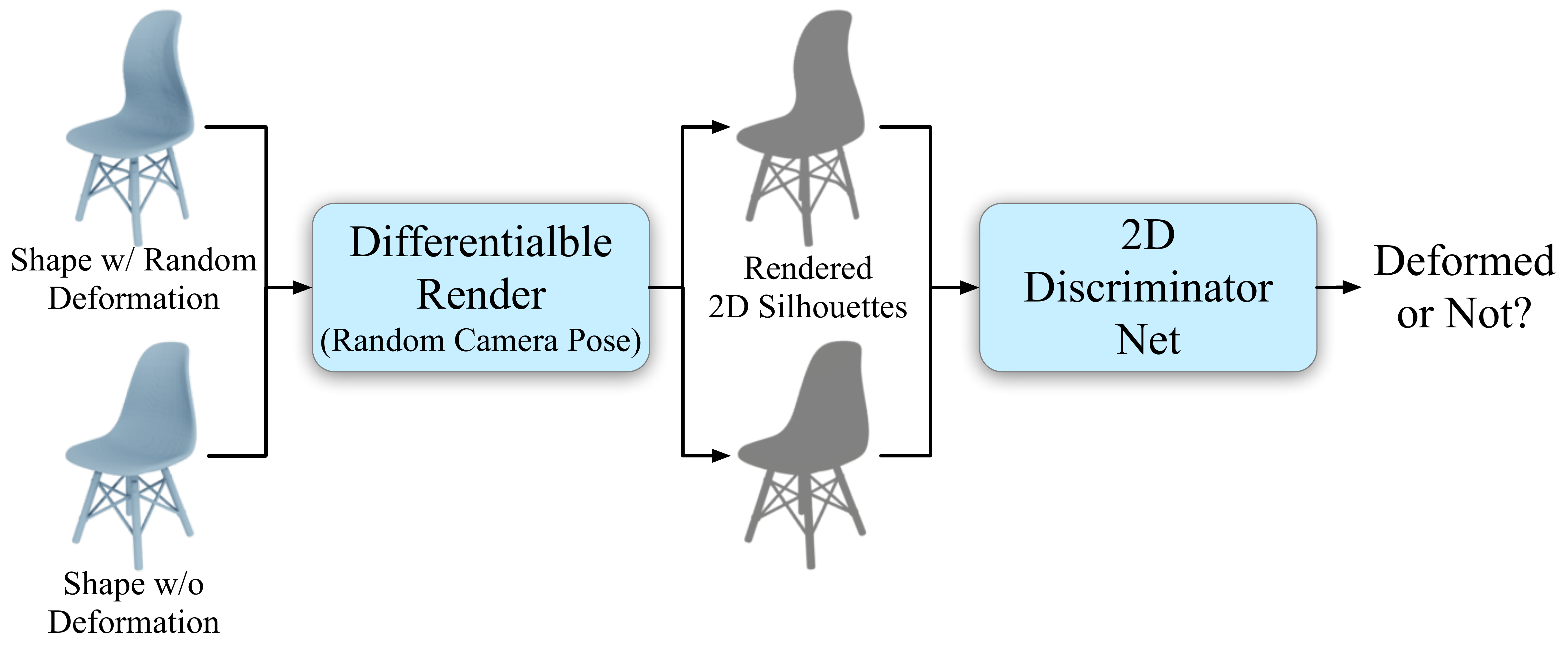}
\end{center}
\vspace{-\baselineskip}
\caption{We utilize a soft rasterizer~\cite{liu2019soft} and a 2D discriminator network to penalize unrealistic deformations.}
\label{fig:discriminator}
\end{figure}

\subsection{Loss Functions}
\label{sec:loss_functions}
\vspace{-3pt}

We consider three objectives when training our network:
1) the deformed input (source) shape matches the given target shape; 2) any deformation sampled from the learned ranges is plausible; 3) the learned meta-handles properly disentangle the deformation space.
We thus train our network with the following joint loss function:
\vspace{-2mm}
\begin{equation}
\mathcal{L} = \mathcal{L}_{fit} + \mathcal{L}_{geo} +  \mathcal{L}_{adv} + \mathcal{L}_{disen}.
\vspace{-2mm}
\end{equation}

Among the four terms, the fitting loss  $\mathcal{L}_{fit}$ corresponds to the first objective and minimizes the Chamfer distance~\cite{Fan:2017} between the deformed source point cloud and the target point cloud.

$\mathcal{L}_{geo}$ and $\mathcal{L}_{adv}$ are geometry loss and adversarial loss, respectively, added for the second objective. In each iteration, we randomly sample a deformation within the predicted ranges, and apply these two losses to penalize implausible deformations. 

Specifically, $\mathcal{L}_{geo}$ is further decomposed into:
\vspace{-2mm}
\begin{equation}
    \mathcal{L}_{geo} =  \mathcal{L}_{symm} + \mathcal{L}_{nor} + \mathcal{L}_{Lap},
\vspace{-2mm}
\label{eq:loss_geo}
\end{equation}
where $\mathcal{L}_{symm}$ is symmetry loss minimizing the Chamfer distance~\cite{Fan:2017} between the deformed point cloud and its reflection along the x-axis (also used in previous works~\cite{3DN,NeuralCages}). Given the mesh connectivity, normal loss $\mathcal{L}_{nor}$ and Laplacian loss $\mathcal{L}_{Lap}$ are computed to prevent distortions. $\mathcal{L}_{nor}$ minimizes the angle difference between the face normals of the source mesh and the deformed mesh. $\mathcal{L}_{Lap}$ minimizes $l1$-norm of the difference of Cotangent Laplacian.

It is not enough to guarantee plausible deformation with only geometric regularization. We thus leverage an adversarial loss $\mathcal{L}_{adv}$, which is defined with a soft rasterizer and a 2D discriminator. (A similar adversarial training idea using 2D projection is also introduced by Li~\etal~\cite{li2019synthesizing}.) As shown in Fig.~\ref{fig:discriminator}, we feed both randomly deformed shapes and shapes without deformation into a soft rasterizer~\cite{liu2019soft}. The renderer captures a soft silhouette image for each shape from a random view. The images are then fed into a simple 2D convolution neural network to predict whether they come from a deformed shape or not. The 2D discriminator network is jointly trained with MetaHandleNet and DeformNet with a classification loss function. The output probabilities for deformed shapes are used to penalize implausible deformations.

For the third objective, we introduce a disentanglement loss $\mathcal{L}_{disen}$. Inspired by Aumentado-Armstrong~\etal~\cite{Armstrong:2019}, $\mathcal{L}_{disen}$ is defined with four terms:
\vspace{-2mm}
\begin{equation}
    \mathcal{L}_{disen} = \mathcal{L}_{sp} +  \mathcal{L}_{cov} + \mathcal{L}_{ortho} + \mathcal{L}_{SVD}.
\vspace{-2mm}
\label{eq:loss_disen}
\end{equation}

Specifically, $\mathcal{L}_{sp}$ encourages the meta-handles $\mathbf{M}_i$ and the coefficient vector $\mathbf{a}$ to be sparse by penalizing their $l1$-norm. $\mathcal{L}_{cov}$ penalizes the covariance matrix (calculated for each batch) of the coefficients $\mathbf{a}$. $\mathcal{L}_{ortho}$ encourages meta-handles to cover different parts of the control-point offsets by penalizing ``dot products'' between the meta-handles.  $\mathcal{L}_{SVD}$ encourages the control points to translate in a single direction within each meta-handle. Please refer to the supplementary materials for the details of $\mathcal{L}_{disen}$. 

Note that we do not incorporate any explicit loss function for the coefficient ranges. While $\mathcal{L}_{fit}$ motivates the coefficient ranges to expand to cover more plausible deformations, $\mathcal{L}_{geo}$ and $\mathcal{L}_{adv}$ prevent the ranges from excessive expansion by penalizing implausible deformations. The coefficient ranges are thus motivated to learn a trade-off.

%% file: sections/results.tex
\section{Experiments}
\vspace{-3pt}

\subsection{Target-Driven Deformation}
\vspace{-3pt}

\begin{figure*}[t]
\begin{center}
   \includegraphics[width=\linewidth]{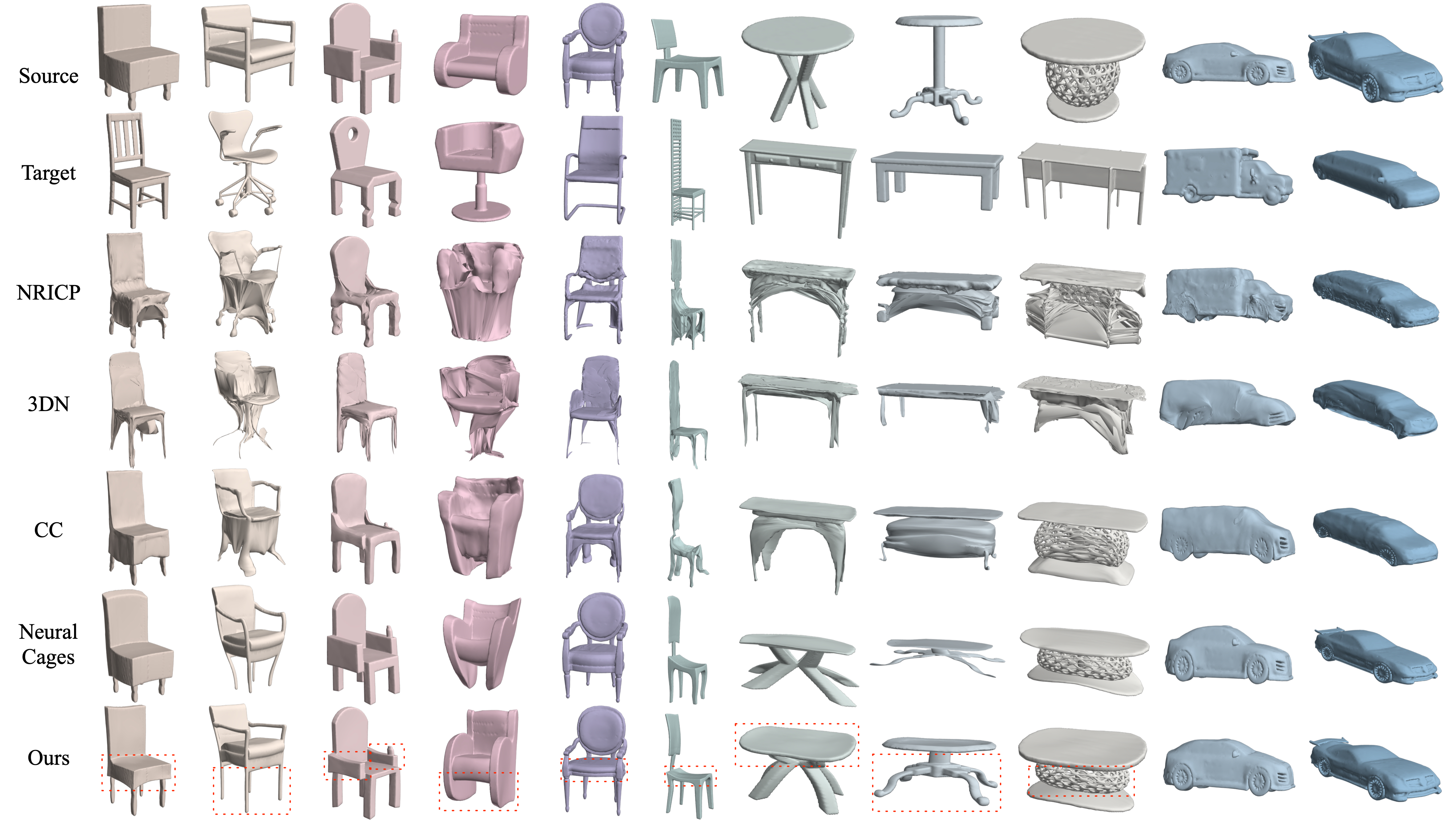}
   \vspace{-1.5\baselineskip}
   \caption{Qualitative comparison of our method with other deformation methods~\cite{huang2017learning,3DN,groueix2019unsupervised,NeuralCages}. Our method allows flexible deformation and fine-grained detail preservation. Our results are also more plausible, especially when the source-target pairs do not share the same structures (see the second and the fourth columns). Please zoom in for details.}
   \label{fig:pair-qualitative}
\end{center}
\vspace{-1.3\baselineskip}
\end{figure*}

We evaluate our methods on the ShapeNet dataset~\cite{ShapeNet}. We choose 15,522 models from the dataset, which cover three categories: chair, table, and car. Shapes are normalized to fit in a unit sphere. For each shape, we sample $c=50$ control-point handles by FPS, in order to generally cover most of the surface and allow flexible deformations. We uniformly sample point clouds of the size $p=4096$ to represent the shapes. We set the number of meta-handles to be $m=15$. This should be an upper bound since the network can use part of them by setting some ranges to zero. As tetrahedral meshes are required as input to compute the biharmonic weights~\cite{wang2015linear}, all the ShapeNet~\cite{ShapeNet} triangular meshes are first fed into Huang~\etal's algorithm~\cite{huang2018robust} to become watertight manifolds, and are then fed into TetWild~\cite{Hu:2018:TMW:3197517.3201353} to produce tetrahedral meshes. We use libigl's~\cite{libigl} implementation to compute the biharmonic coordinates, which are then interpolated from the mesh vertices to the sampled point cloud. For the differentiable renderer, we use an implementation from Pytorch3D~\cite{ravi2020accelerating}. We reserve $10\%$ models for testing and the rest for training. For each category, we train a separate model and test it on $3,000$ randomly sampled source-target pairs.

We compare our method to non-rigid ICP (NRICP)~\cite{huang2017learning}, a non-neural registration technique which aligns two point clouds by minimizing a smooth deformation energy; 3D deformation network (3DN)~\cite{3DN} and cycle‐consistent deformation (CC)~\cite{groueix2019unsupervised}, two learning-based methods that directly infer per-vertex displacements; and Neural Cages~\cite{NeuralCages}, a learnable cage-based deformation method. 

\begin{figure}[t]
\begin{center}
   \includegraphics[width=\linewidth]{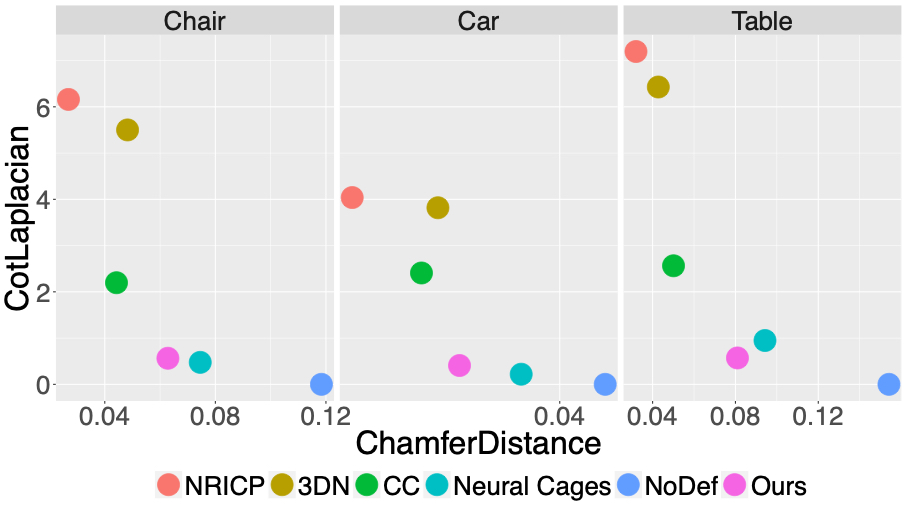}
   \vspace{-1\baselineskip}
   \caption{Quantitative comparison of the target-driven deformation. Each 2D point represents one method. The coordinates correspond to the alignment error and the distortion, with the origin being ideal. `NoDef' indicates undeformed source shapes.}
   \label{fig:pair-quantitative}
\end{center}
\vspace{-1\baselineskip}
\end{figure}

Qualitative results are shown in Fig.~\ref{fig:pair-qualitative}. Although NRICP~\cite{huang2017learning}, 3DN~\cite{3DN}, and CC~\cite{groueix2019unsupervised} do align the source shape to the target shape in most cases, they fail to preserve fine-grained details of the source shape and introduce lots of distortions. The results of Neural Cages~\cite{NeuralCages} look more pleasing, but the cage-based deformation is less flexible than our control-point based deformation. Compared to the Neural Cages~\cite{NeuralCages}, our method can achieve more detailed deformation of a local region, such as adjusting the thickness of chair seats (first and fifth columns) and armrests' height (third column). Also, most alternative methods produce unrealistic deformations when the source shape and the target shape do not share similar structures. For example, suppose the source shape has four chair legs, and the target shape is a swivel chair (second and fourth columns). In that case, the alternative methods tend to deform the four chair legs toward the center under the fitting loss's influence, resulting in undesirable deformations. Thanks to the adversarial regularization we employed, our method can avoid such implausible deformations while still aligning the output to the target.

\begin{figure}[t]
\begin{center}
   \includegraphics[width=\linewidth]{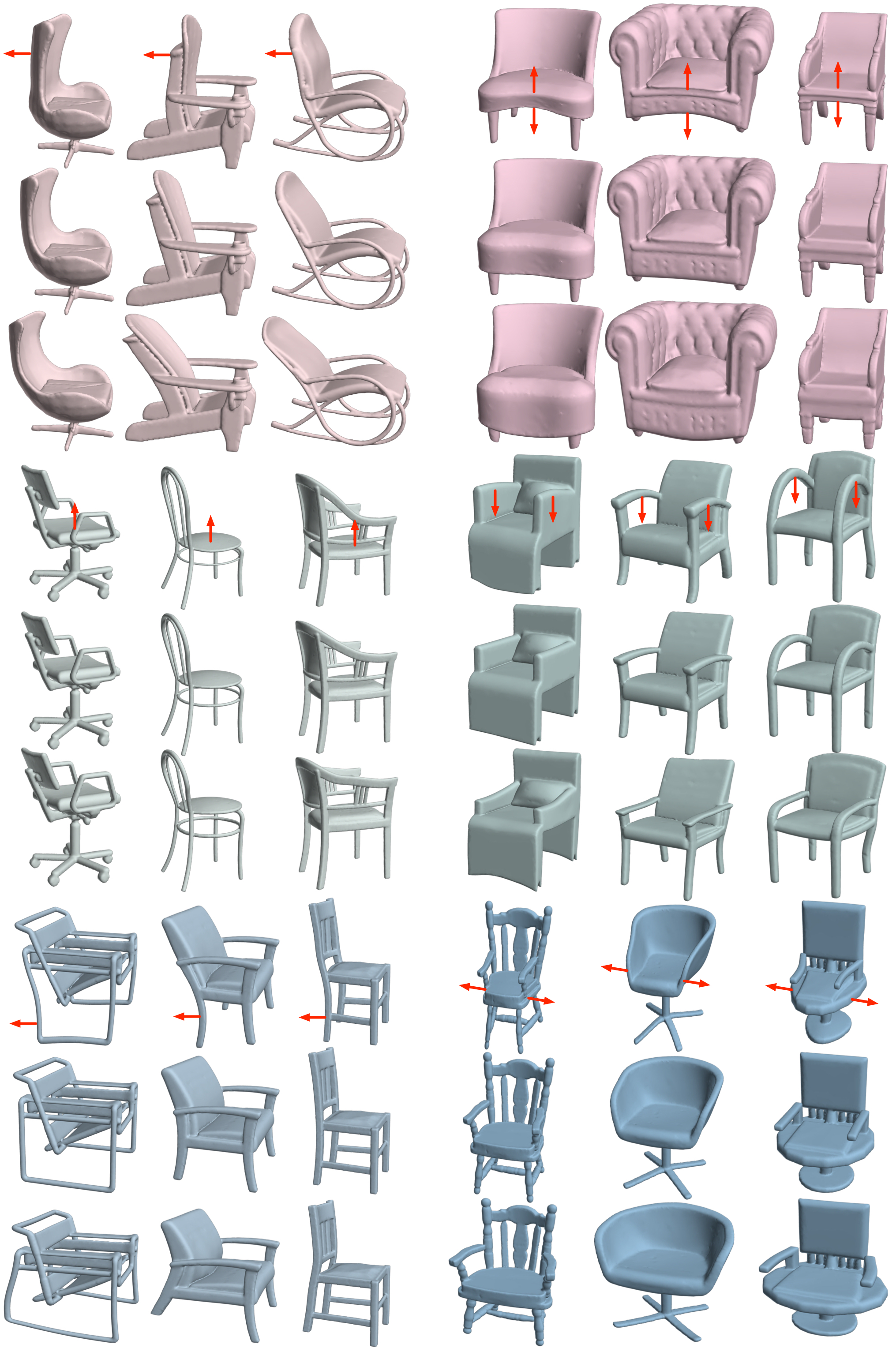}
   \vspace{-1.5\baselineskip}
   \caption{Learned meta-handles across different shapes. The figure includes six meta-handles, and each color indicates a distinct one. For each meta-handle, the figure demonstrates the corresponding deformations on three different shapes, with the red arrows highlighting the deformation direction. The meta-handles are consistent across various shapes.}
   \label{fig:multiple-shape-meta-handle}
\end{center}
\vspace{-1\baselineskip}
\end{figure}

Inspired by Neural Cages~\cite{NeuralCages}, we also utilize Chamfer distance~\cite{Fan:2017} between the deformed shape and the target shape (computed over 100,000 uniformly sampled points) to measure the alignment error; and use the difference between cotangent Laplacians of the source shape and the deformed shape ($l1$-norm) to measure the distortion. The quantitative results are shown in Fig.~\ref{fig:pair-quantitative}. As shown in the figure, although NRICP~\cite{huang2017learning}, 3DN~\cite{3DN}, and CC~\cite{groueix2019unsupervised} achieve lower alignment errors, the distortions are much higher. Compared to Neural Cages~\cite{NeuralCages}, our method achieves better Chamfer distance with a similar cotangent Laplacian.

\subsection{Meta-Handle Deformation Space}
\vspace{-3pt}

Another main contribution of our method is that, for each shape, we learn a set of interpretable meta-handles with the corresponding coefficient ranges, which factorize all the plausible deformations for the shape.

\input{tables/comparison_plausibility}

Fig.~\ref{fig:teaser} demonstrates some learned meta-handles of a single shape. Each column shows the deformations along the direction of a meta-handle, with the deformation scale uniformly sampled within the corresponding coefficient range. The red arrows highlight the deformation direction of each meta-handle. As shown in the figure, the learned meta-handles are disentangled and factorize all the plausible deformations for the shape. Although we do not take any semantic annotation or correspondences across different shapes as input or supervision, our method is able to learn some intuitive meta-handles. Specifically, the learned meta-handles are not limited to global scaling. Many of them align with some local semantic parts, such as adjusting the thickness of the chair seat (first column), the height of armrests (fourth column), the length of four chair legs (seventh column), and the height of the chair back (eighth column). Also, many of them involve non-rigid deformation of some parts, such as bending the chair back (fifth column) and two back legs (sixth column), which cannot be achieved through the rigid bounding-box handles proposed by previous methods~\cite{gadelha2020learning,Sung:2020}. To construct a low-dimensional compact deformation space, the learned meta-handles not only leverage correlations between the control-point handles, but also discover the underlying hard constraints (e.g., symmetry) of the shape structure. Meanwhile, the coefficient ranges learn the underlying soft priors (e.g., ratios of part scales) and provide reasonable deformation scopes for meta-handles.

We assume that, for different shapes, meta-handles with the same index share similar deformations due to the structure feature of MetaHandleNet. As shown in Fig.~\ref{fig:multiple-shape-meta-handle}, our learned meta-handles are consistent across different shapes. Despite geometry details and even global structures being different, each meta-handle can find corresponding regions across various shapes and predict similar deformations, which is interesting as we do not provide any semantic annotation or correspondence information.

Inspired by Achlioptas~\etal~\cite{Achlioptas:2018}, we also employ coverage (COV) and minimum matching distance (MMD) to evaluate our generative model. For a set of generated shapes $A$ and a set of ground truth shapes $B$, coverage measures the \emph{fraction} of the shapes in $B$ that can be roughly represented within $A$, while MMD measures \emph{how well} shapes in $B$ can be represented by shapes in $A$. For both metrics, closeness is computed using Chamfer distance~\cite{Fan:2017}. For each category, we separate 500 shapes for constructing the set $A$, and the remaining shapes are regarded as set $B$. For our method, we randomly sample 20 deformations within the learned deformation space of each shape. For baseline methods 3DN~\cite{3DN}, CC~\cite{groueix2019unsupervised}, and Neural Cages~\cite{NeuralCages}, we randomly sample 20 target shapes for each shape in $A$ to generate target-driven deformations. The quantitative results are shown in Table~\ref{tab:variation-comparison}. While all the methods have similar MMDs, our method achieves higher coverages, which indicates that our method generates more diverse deformations, and more ground truth shapes can thus be represented within our deformation space.

\subsection{Ablation Studies}
\vspace{-3pt}

\input{tables/ablation_fitting}

\paragraph{Meta-Handles} Instead of predicting a set of meta-handles, we can deform a shape by directly predicting the offset of each control-point handle (deformation function $f$). We compare our method to this variant. As shown in Table~\ref{tab:ablation-fitting}, when there is no adversarial loss (first and second rows), directly using $50$ control-point handles can achieve better fitting error and smaller distortion, since it allows more degrees of freedom for the deformation. However, when applied with the adversarial loss (third and fourth rows), it is harder for the network to find plausible deformations based on 50 control-point handles, while our learned meta-handles provide intuitive deformations resulting in better results.
\begingroup
\setlength{\columnsep}{3mm}
\begin{wraptable}{r}{0.5\linewidth}
\footnotesize
\setlength{\tabcolsep}{1pt}
\vspace{-3mm}
  \caption{Coverage (higher is better) and MMD ($\times 100$, lower is better) for different ablated versions (on Chair category).}
  \vspace{-3mm}
    \begin{tabularx}{\linewidth}{c|C|C}
    \toprule
          & COV$\uparrow$   & MMD$\downarrow$ \\
    \hline
    w/o meta-handle & 48.4\% & 4.69 \\
    \hline
    w/o $\mathcal{L}_{adv}$ & 56.3\% & 4.64 \\
    \hline

    w/o $\mathcal{L}_{disen}$ & 64.1\% & 4.14 \\
    \hline
    Ours  & 64.6\% & 4.28 \\
    \bottomrule
    \end{tabularx}%
    \vspace{-1\intextsep}
  \label{tab:ablation_variation}%
\end{wraptable}
  Also, without meta-handles, we cannot directly sample plausible variants of a input shape. The target-driven deformation is less effective in generating diverse deformations and covering all the plausible variants (see the first row of Table~\ref{tab:ablation_variation}). 

\endgroup 
\vspace{-1.2em} 
\begingroup
\setlength{\columnsep}{3mm}%
\begin{wrapfigure}{r}{0.5\linewidth}\vspace{-5mm}
\includegraphics[width=\linewidth]{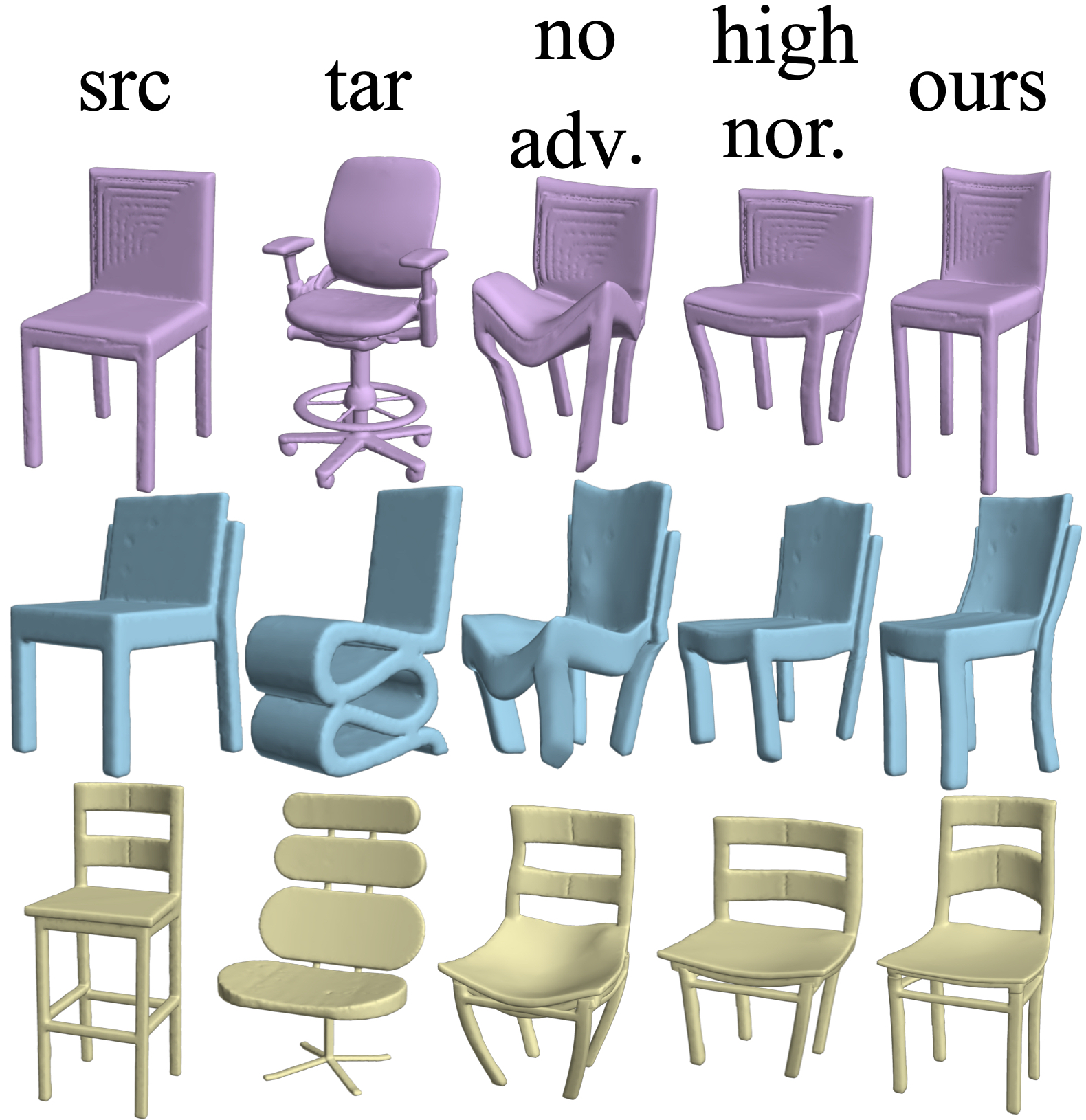}
\vspace{-7mm}
\caption{Comparison between $\mathcal{L}_{nor}$ and $\mathcal{L}_{adv}$. Both the third column and the fourth column have no $\mathcal{L}_{adv}$, but the fourth column has higher weight for $\mathcal{L}_{nor}$.}\label{fig:normal_adv}
\vspace{-3mm}
\end{wrapfigure}
\paragraph{Adversarial Regularization} We use both adversarial loss $\mathcal{L}_{adv}$ and normal loss $\mathcal{L}_{nor}$ (part of the geometric loss $\mathcal{L}_{geo}$) to encourage plausible deformations. Fig.~\ref{fig:normal_adv} demonstrates a qualitative comparison between them. When there is no $\mathcal{L}_{adv}$, the deformation may lose plausibility in order to match the target shape. Although $\mathcal{L}_{nor}$ can also alleviate this issue to some extent, strong $\mathcal{L}_{nor}$ (fourth column) may be too restrictive for the deformation, while $\mathcal{L}_{adv}$ achieves more realistic results and still allows flexible deformations. When $\mathcal{L}_{adv}$ is applied, the fitting error increases (second and fourth row of Table~\ref{tab:ablation-fitting}) in exchange for more plausible deformations. As shown in Table~\ref{tab:ablation_variation}, without $\mathcal{L}_{adv}$, both the coverage and MMD become worse, indicating that $\mathcal{L}_{adv}$ is important for generating diverse and realistic deformations.

\endgroup 
\vspace{-4mm}
\paragraph{Disentanglement Regularization} 
We use $\mathcal{L}_{disen}$ to encourage the intuitive factorization of the deformation space. As shown in Fig.~\ref{fig:disentangle}, when there is no $\mathcal{L}_{disen}$, the deformations along each learned meta-handle are still plausible, since $\mathcal{L}_{geo}$ and $\mathcal{L}_{adv}$ are still applied to the random samples within the space to penalize unrealistic deformations. However, the learned meta-handles are entangled, each 
\begingroup
\setlength{\columnsep}{3mm}%
\begin{wrapfigure}{r}{0.5\linewidth}\vspace{-3mm}
   \includegraphics[width=\linewidth]{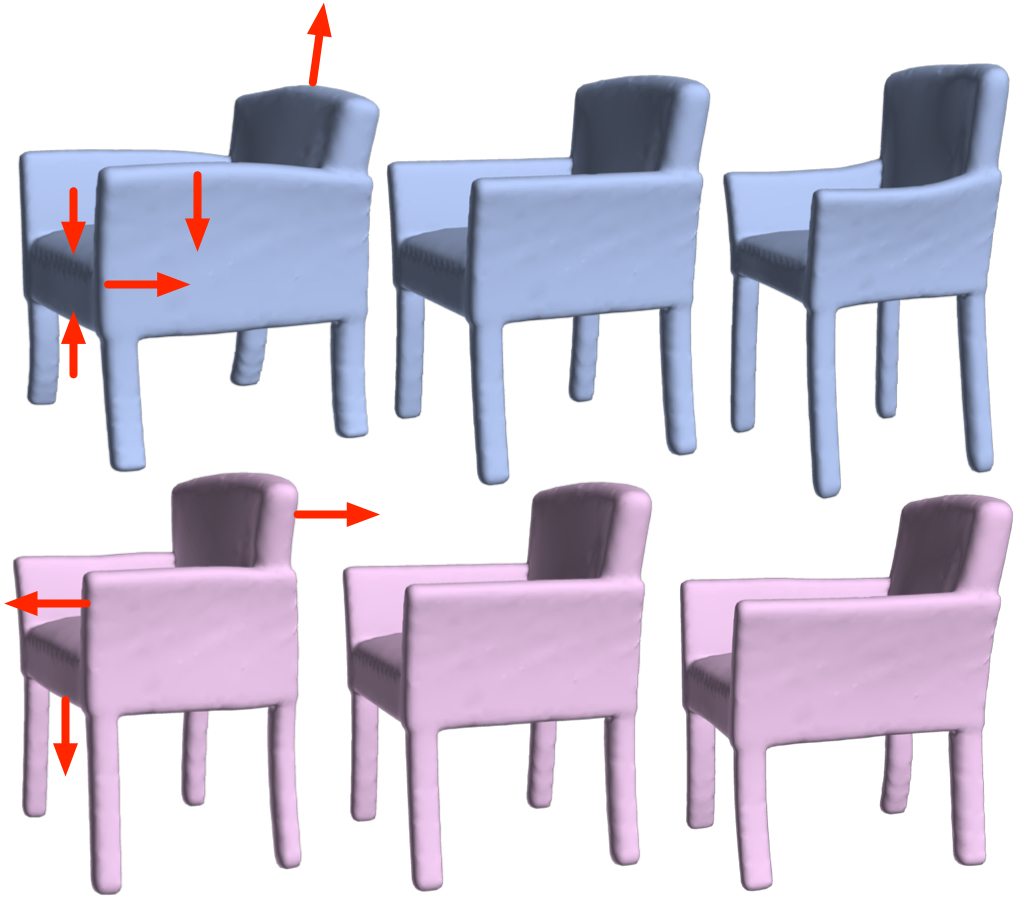}
\vspace{-5mm}
   \caption{Results w/o $\mathcal{L}_{disen}$, each row indicates a learned meta-handle.}
   \label{fig:disentangle}
\vspace{-3mm}
\end{wrapfigure}
 meta-handle may deform multiple parts along different directions, and there are overlappings between different meta-handles. In contrast, meta-handles in Fig.~\ref{fig:teaser}  provide more intuitive and disentangled deformations. Table~\ref{tab:ablation_variation} quantitatively verifies that $\mathcal{L}_{disen}$ does not affect the diversity and plausibility of the deformation space.

\endgroup

%% file: tables/comparison_plausibility.tex
\begin{table}[t]
{
\footnotesize
  \setlength{\tabcolsep}{0.2em}
  \centering
  \caption{Coverage (higher is better) and MMD ($\times 100$, lower is better) comparison between different methods.}
  \vspace{-\baselineskip}
    \begin{tabularx}{\linewidth}{c|C|C|C|C|C|C}
    \toprule
    \multirow{2}[4]{*}{} & \multicolumn{2}{c|}{Chair} & \multicolumn{2}{c|}{Car} & \multicolumn{2}{c}{Table} \\
      \cline{2-7}
          & COV $\uparrow$ & MMD $\downarrow$ & COV $\uparrow$   & MMD $\downarrow$ & COV $\uparrow$   & MMD $\downarrow$ \\
    \midrule
    3DN~\cite{3DN} & 32.0\% & 4.56 & 46.6\% & 2.91 & 30.6\% & 4.26 \\
   CC~\cite{groueix2019unsupervised} & 51.0\% & 4.26 & 50.3\% & 2.79 & 50.2\% & 3.88\\
   NC~\cite{NeuralCages} & 54.4\% & 4.23 & 66.6\% & 2.65 & 44.7\% & 3.85 \\
    Ours  & 64.6\% & 4.28 & 76.5\% & 2.97 & 54.9\% & 3.70 \\
    
    \bottomrule
    \end{tabularx}%
  \label{tab:variation-comparison}%
}
\end{table}%

%% file: tables/ablation_fitting.tex
\begin{table}[t]
\footnotesize
  \centering
  \caption{Chamfer distance ($\times 100$) and Cotangent Laplacian ($\times 10$) between different ablated versions (on chair category). For both metrics, lower is better. DoF indicates degrees of freedom.}
  \vspace{-\baselineskip}
    \begin{tabularx}{\linewidth}{c|C|C|C|C}
    \toprule
    Meta-handle / Handle & DoF   & $\mathcal{L}_{adv}$ & CD$\downarrow$   & CotLap$\downarrow$ \\
    \midrule
    Handle & $50\times3$   & w/o   & 4.78 & 5.60 \\
    Meta-handle  & 15    & w/o   & 5.76 & 8.61 \\
    Handle & $50\times3$   & w/    & 7.98 & 7.69 \\
    Meta-handle  & 15    & w/    & 6.28 & 5.75 \\
    \bottomrule
    \end{tabularx}%
  \label{tab:ablation-fitting}%
\end{table}

%% file: sections/conclusion.tex
\vspace{-0.5mm}
\section{Conclusion}
\vspace{-1.5mm}
We presented \textbf{DeepMetaHandles}, a 3D conditional generative model based on mesh deformation. Our method takes automatically-generated control points with biharmonic coordinates as deformation handles, and learns a latent space of deformation for each input mesh. Each axis of the space is explicitly associated with multiple deformation handles, and it's thus called a meta-handle. The disentangled meta-handles factorize all the plausible deformations of the shape, while each of them conforms to an intuitive deformation. We learn the meta-handles unsupervisely by incorporating a target-driven deformation module. We also employ a differentiable render and a 2D discriminator to enhance the plausibility of the deformation.

In our method, the expressibility of the deformation is limited by the given control points. Technically, increasing the number of input control points a lot will result in a memory issue and making the network training more difficult. An interesting future direction would be developing another network that can adaptively sample the control points at appropriate locations and thus enable more fine-grained local deformations.

\vspace{-3mm}

%% file: sections/supplementary.tex

\ifpaper
  \newcommand\refpaper[1]{\unskip}
\else
  \makeatletter
  \newcommand{\manuallabel}[2]{\def\@currentlabel{#2}\label{#1}}
  \makeatother
  \manuallabel{sec:learning_meta_handles}{3.2}
  \manuallabel{eq:loss_disen}{5}
  \manuallabel{fig:pair-qualitative}{6}

  \newcommand{\refpaper}[1]{in the paper}
\fi

\begin{figure*}[t]
\begin{center}
    \includegraphics[width=\linewidth]{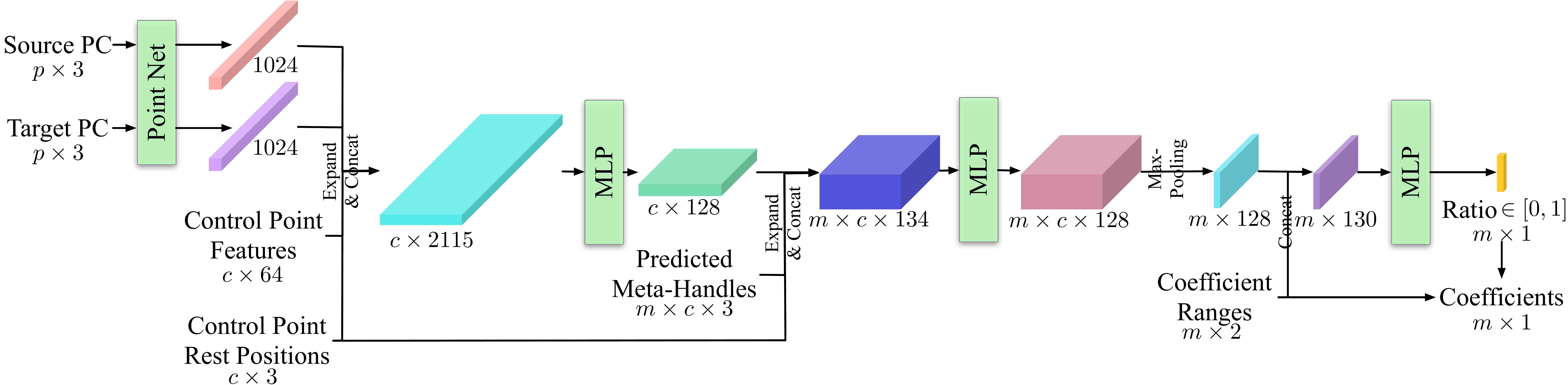}
    \captionof{figure}{Architecture of DeformNet.}
  \label{fig:deformnet}
\end{center}
\end{figure*}

Please check out our webpage\footnote{\url{http://cseweb.ucsd.edu/~mil070/deep_meta_handles_supp_animations}} for the animations of the learned meta-handles. In this supplementary material, we first discuss the network architecture details of the range prediction module, DeformNet, and the discriminator network (Sec.~\ref{sec:supp_architecture}), and also the training details (Sec.~\ref{sec:training_details}). We then present the details and ablation studies of the disentanglement loss $\mathcal{L}_{disen}$ (Sec.~\ref{sec:supp_disen_loss}). Moreover, we evaluate the impact of the numbers of control points and the output meta-handles (Sec.~\ref{sec:supp_num_control_points}) and also discuss the difference between our differentiable-renderer-based 2D discriminator and 3D discriminator (Sec.~\ref{sec:3d_discriminator}). We also examine the effectiveness of our deformation generative model when it is used for data augmentation (Sec.~\ref{sec:supp_data_aug}). Lastly, we provide more results of the target-driven deformation and show learned meta-handles for the laptop category (Sec.~\ref{sec:supp_more_target_driven}).

\subsection{Animations of the Learned Meta-Handles}
\label{sec:supp_animations}

We show the animations of the learned meta-handles in our webpage. Chrome browser is preferred for the best display. On the webpage, each row shows deformations of meta-handles with the \emph{same} index for different shapes. Note that the learned meta-handles are \emph{consistent} across the shapes. The animations demonstrate that our learned meta-handles properly factorize the plausible deformation space of the shape while each of them corresponds to an intuitive deformation direction.

\subsection{Network Architecture}
\label{sec:supp_architecture}
\begin{figure}[t]
\begin{center}
   \includegraphics[width=\linewidth]{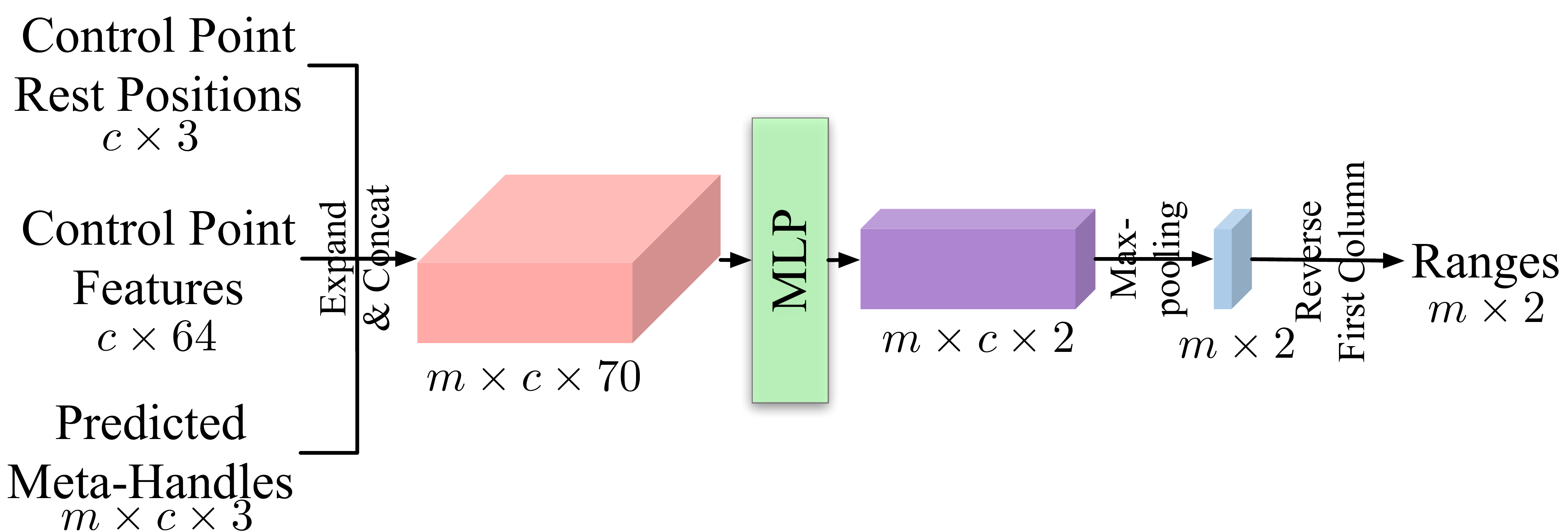}
   \caption{Architecture of the range prediction module.}
   \label{fig:range_module}
\end{center}
\end{figure}

In this subsection, we describe the architectures of the range prediction module, DeformNet, and the discriminator network, which are introduced in Sec.~\ref{sec:learning_meta_handles}~\refpaper~.

\vspace{-1em}
\paragraph{Range Module} As shown in Fig.~\ref{fig:range_module}, after predicting the meta-handles, the range module predicts a coefficient range $[L_i, R_i]$ for each meta-handle. It takes the rest positions of the control points, 64-dimensional control point features (predicted by MetaHandleNet), and the predicted meta-handles as input. The module incorporates the information by building a 3D tensor, where each pair of meta-handle and control point has a 70-dimensional feature. The module then applies an MLP to the 70-dimensional features, resulting in a 2-dimensional feature for each pair of meta-handle and control point. The module then utilizes a max-pooling to aggregate the information across all the control points, resulting in a $m \times 2$ matrix. We then reverse the sign of the first column (due to the max-pooling) to output the final coefficient ranges.
\vspace{-1em}
\paragraph{DeformNet} After MetaHandleNet predicts a set of meta-handles with the corresponding coefficient ranges for the source shape, DeformNet finds a coefficient vector within the deformation space so that the deformed source shape matches the target shape. Fig.~\ref{fig:deformnet} shows the architecture of DeformNet. It first utilizes PointNet~\cite{PointNet} to process both the source and target point clouds to obtain the global features for the source and target shapes. The global features are repeated for the control points and are then combined with the 64-dimensional control point features (predicted by MetaHandleNet) and the rest positions of the control points, resulting in a 2,115-dimensional feature for each control point. The features are fed into an MLP to output a 128-dimensional feature for each control point. We then create a 3D tensor to incorporate the control point features, predicted meta-handles, and the rest positions of the control points. In this 3D tensor, each pair of meta-handle and control point has a 134-dimensional feature. We then apply an MLP to the features to output a 128-dimensional feature for each pair of meta-handle and control point. A max-pooling is then applied to aggregate the features across all the control points, resulting in a 128-dimensional feature for each meta-handle. The features are then combined with the predicted coefficient ranges and are fed into another MLP (with Sigmoid as the final activation function) to output a ratio within $[0,1]$ for each meta-handle. With both the ratios and the coefficient ranges, we output a coefficient vector within the ranges to represent the deformation.
\vspace{-1em}
\paragraph{Discriminator} We utilize a relatively simple 2D network as the discriminator network to match the capability of the deformation generation part. Specifically, the discriminator network takes a $128\times128$ image as input and uses three convolutional layers to process. Each convolutional layer is followed by batch normalization and LeakyReLU. A fully connected layer, along with the sigmoid function, is then utilized to output the final probability.

\subsection{Training Details}
\label{sec:training_details}

As described in Sec.\ 4.1, we use public code to convert the mesh to tetrahedral mesh and then calculate the biharmonic coordinates. We found that the mesh conversion and the coordinate computation are robust even to the shapes with thin parts and complicated topology. Note that we have an example with a complex wire structure in the 9th column of Fig.\ 6. In the network training, however, we also find that pruning some shapes that have large biharmonic coordinates is helpful for faster convergence. We removed 10\% of such shapes in the ShapeNet dataset.

We trained our models on 3 Nvidia RTX 2080 Ti GPUs for $2.8 \times 10^4$ iterations (i.e., $1.1 \times 10^6$ pairs) with a batch size of 39. Adam is used as the optimizer with a learning rate of 1e-4. All loss terms have an individual weight, and we empirically select the weights. For chair category, the weights are set to 1, 1, 0.1, 3, 6e-3, 1e-3, 1e-3, 1e-3, and 0.3 for $\mathcal{L}_{fit}$, $\mathcal{L}_{symm}$, $\mathcal{L}_{nor}$, $\mathcal{L}_{Lap}$, $\mathcal{L}_{adv}$, $\mathcal{L}_{sp}$, $\mathcal{L}_{cov}$, $\mathcal{L}_{ortho}$, and $\mathcal{L}_{SVD}$ respectively.

\subsection{Details of the Disentanglement Loss $\mathcal{L}_{disen}$}
\label{sec:supp_disen_loss}
Here, we introduce the four terms of the disentanglement loss $\mathcal{L}_{disen}$ (Eq.~\ref{eq:loss_disen}~\refpaper~).

$\mathcal{L}_{sp}$ is a sparsity loss that encourages the meta-handles $\mathbf{M}_i$ and the coefficient vector $\mathbf{a}$ to be sparse by penalizing their $l1$-norm:
\begin{equation}
    \mathcal{L}_{sp} = \frac{1}{m}\sum_{i=1}^{m}\| \mathbf{M}_i\|_1 + \| \mathbf{a}\|_1,
\end{equation}
where $m$ is the number of the meta-handles.

$\mathcal{L}_{cov}$ is a covariance penalty loss introduced by Aumentado-Armstrong~\etal~\cite{Armstrong:2019} that encourages meta-handles to be independent with each other. This loss calculates the covariance matrix of the coefficients $\mathbf{a}$ for each batch and penalizes the $l1$-norm of the matrix:
\begin{equation}
    \mathcal{L}_{cov} =  \|\operatorname{cov}(\mathbf{a}, \mathbf{a})\|_1 .
\end{equation}

$\mathcal{L}_{ortho}$ is an orthogonality loss that encourages the meta-handles to cover different coordinates of control point offsets. It is calculated as:
\begin{equation}
    \mathcal{L}_{ortho} =  \sqrt{\sum_{i\neq j} \|\mathbf{M}_i \circ \mathbf{M}_j\|_{1,1}^2 },
\end{equation}
where `$\circ$' denotes element-wise multiplication. Intuitively, if two meta-handles have no overlap over the offset coordinates, we regard them to be ``orthogonal'' and they have zero contribution to $\mathcal{L}_{ortho}$.

Lastly, $\mathcal{L}_{SVD}$ is an SVD loss that encourages the control points to translate along with similar directions within each meta-handle. Specifically, for each meta-handle $\mathbf{M}_i$, we regard the control-point offsets of $\mathbf{M}_i$ as $c$ points in the 3D space. Given the points, we find the best-fit plane and then calculate the sum of the distances from the points to the plane, which is equal to $\sigma_{3}(\mathbf{M}_i^T\mathbf{M}_i)$ and $\sigma_{3}$ indicates the smallest singular value of the matrix. $\mathcal{L}_{SVD}$ is defined as minimizing the distances:
\begin{equation}
    \mathcal{L}_{SVD} =  \frac{1}{m}\sum_{i=1}^{m}\sigma_{3}(\mathbf{M}_i^T\mathbf{M}_i).
\end{equation}

\begin{table}[t]
{\small
  \centering
  \caption{Quantitative comparison between without and with the disentanglement loss $\mathcal{L}_{disen}$. The network is trained on the chair category.}
    \begin{tabularx}{\linewidth}{C|C|C}
    \toprule
          & \multicolumn{1}{c|}{w/o $\mathcal{L}_{disen}$} & \multicolumn{1}{c}{w/ $\mathcal{L}_{disen}$} \\
    \midrule
    $\mathcal{L}_{sp} \downarrow$ & 13.355 & 7.7468 \\ \hline
    $\mathcal{L}_{cov} \downarrow$ & 3.5890 & 1.3229 \\ \hline
    $\mathcal{L}_{ortho} \downarrow$ & 10.843 & 4.2842 \\ \hline
    $\mathcal{L}_{SVD} \downarrow$ & 0.1063 & 0.0002 \\
    \bottomrule
    \end{tabularx}%
  \label{tab:disen}%
}
\end{table}%

\begin{figure}[t]
\begin{center}
   \includegraphics[width=\linewidth]{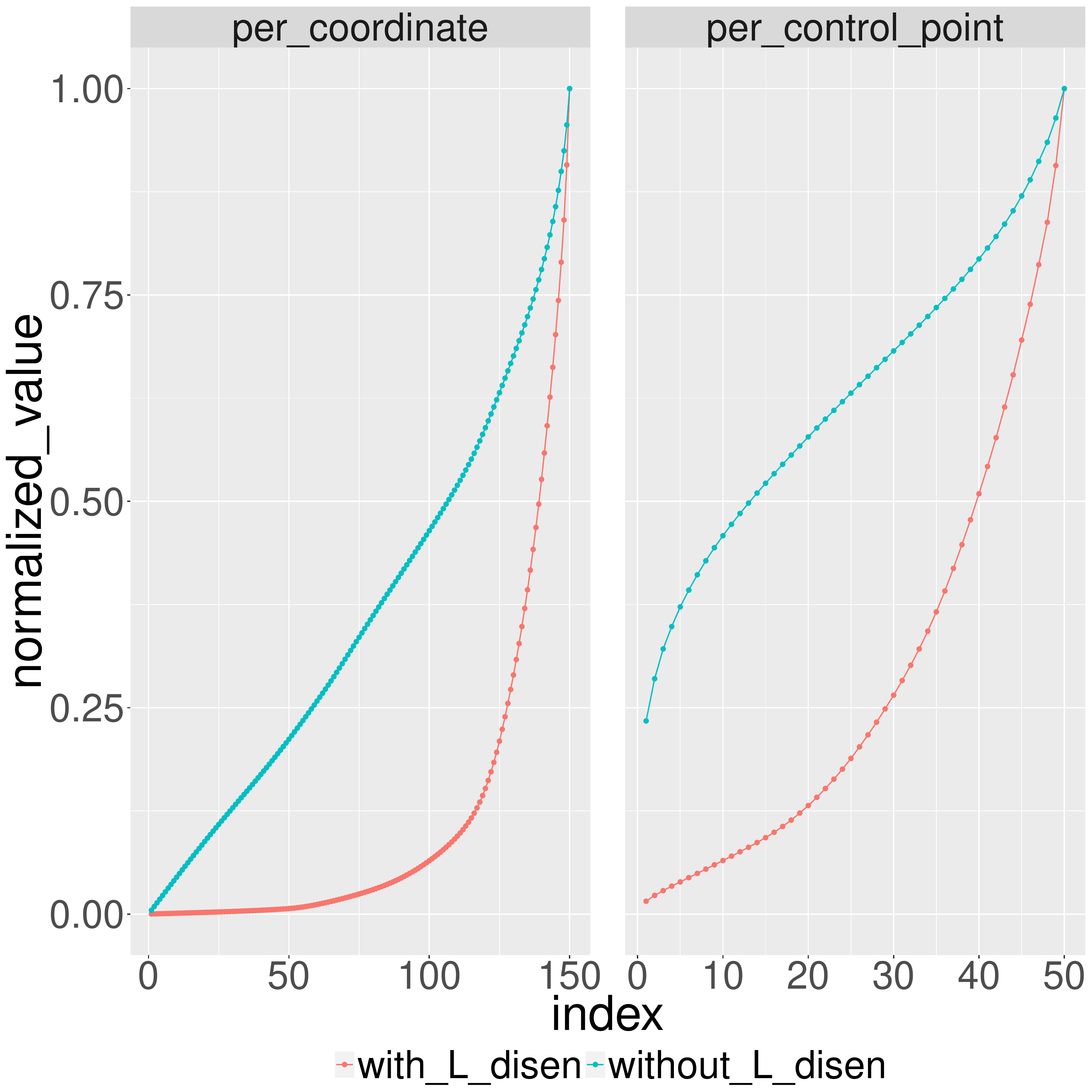}
   \caption{The impact of $\mathcal{L}_{disen}$ on the sparsity of meta-handles.}
   \label{fig:sparsity}
\end{center}
\end{figure}

\begin{figure}[t]
\begin{center}
   \includegraphics[width=\linewidth]{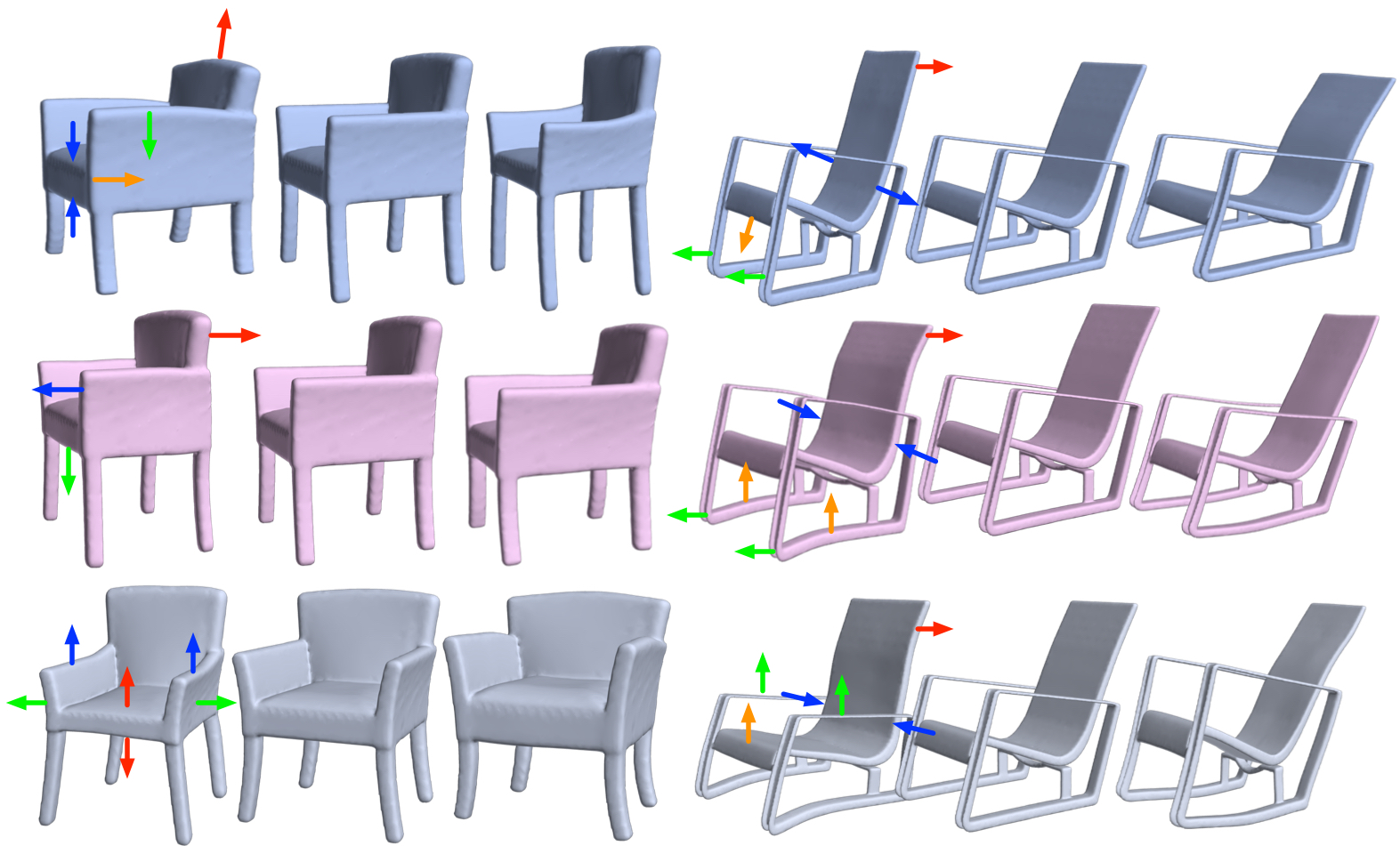}
   \caption{Results without $\mathcal{L}_{disen}$: the figure shows six meta-handles of two shapes, with arrows highlight the deformations. Each meta-handle corresponds to multiple deformations, and there are overlapping between the meta-handles.}
   \label{fig:sup_disentangle}
\end{center}
\end{figure}

Table~\ref{tab:disen} shows the quantitative comparison for the disentanglement loss $\mathcal{L}_{disen}$. We find that after removing $\mathcal{L}_{disen}$, all the four terms increase a lot, which indicates that $\mathcal{L}_{disen}$ is essential for the proper factorization of the deformation space. 

Fig.~\ref{fig:sparsity} further illustrates the impact of $\mathcal{L}_{disen}$ on the sparsity of meta-handles. For the per-coordinate case, we show the distribution of offsets of $50\times 3$ coordinates. For the per-control-point case, we first calculate the $l2$-norm of the offsets for each control point and then show the distribution of the $l2$-norms. For both cases, the values are normalized within each meta-handle and averaged across all the meta-handles and shapes. As shown in the figure, when $\mathcal{L}_{disen}$ is applied, each meta-handle tends to be sparse, and only a small part of coordinates (control points) are impacted. When $\mathcal{L}_{disen}$ is not applied, however, the meta-handles are no longer sparse and tend to deform most of the control points.

Fig.~\ref{fig:sup_disentangle} shows more results when $\mathcal{L}_{disen}$ is ablated. The results demonstrate the importance of $\mathcal{L}_{disen}$ for the proper factorization of the deformation space.

\subsection{Impact of the Numbers of Control Points and Meta-Handles}
\label{sec:supp_num_control_points}

\begin{figure}[t]
\begin{center}
   \includegraphics[width=0.8\linewidth]{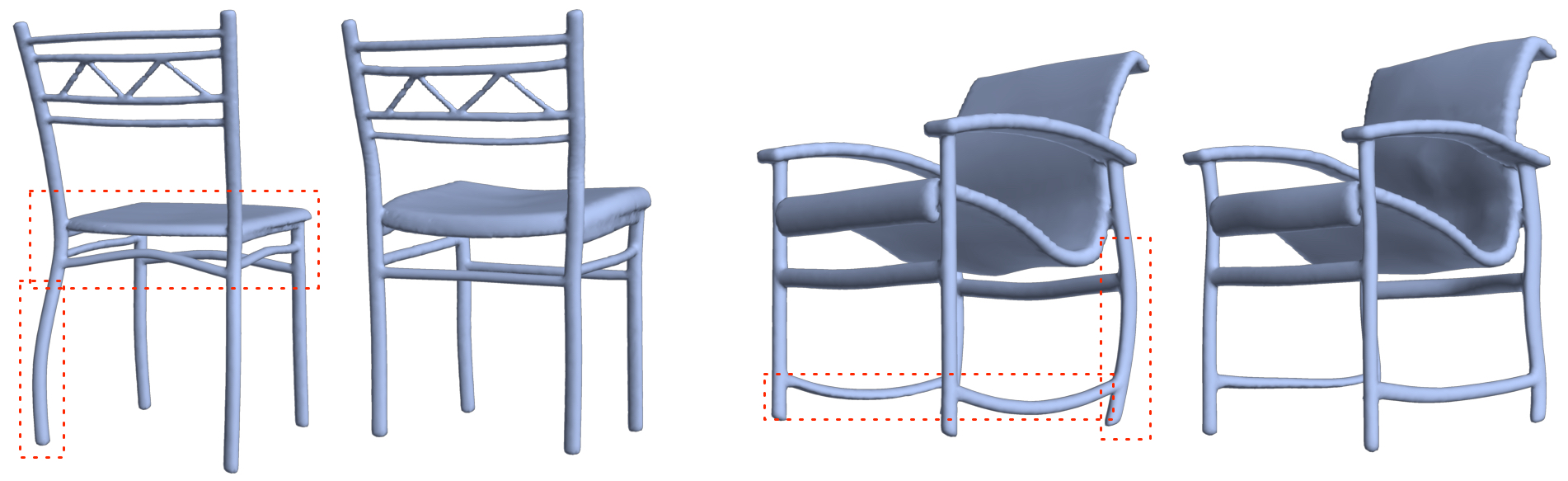}
   \caption{In each pair, the left one shows the deformation with a 3D point cloud discriminator, while the right one shows the deformation with our 2D discriminator.}
   \label{fig:3dgan}
\end{center}
\end{figure}

We evaluate the impact of the number of control points and meta-handles on the deformation. Please note that here the number of meta-handles indicates the upper bound of the size since the network can use part of them by setting the ranges to zero. Table~\ref{tab:control_point} shows the quantitative results of the chair category. As shown in the table, when we decrease the number of control-point handles from 50 to 25, both the Chamfer distance and cotangent Laplacian increase a little bit, since the degree of freedom of the deformation drops. However, when we increase the number of control-point handles from 50 to 100 and 200, both the Chamfer distance and cotangent Laplacian increase a lot. There are two possible reasons: (a) it is more difficult for the network to deform shapes with such a large number of control points; (b) due to the GPU memory bottleneck, we have to reduce the batch size during training when there is a large number of control points (we reduced the batch size from 39 to 12 when increasing the number of control points from 50 to 200). As for the meta-handles, when we double the number of meta-handles, the Chamfer distance is similar, which indicates that 15 meta-handles are already enough to produce flexible deformations for the chair category. However, the cotangent Laplacian becomes worse, which suggests that it is more difficult for the network to handle lots of meta-handles, and they may introduce some unnecessary distortions.

\begin{table}[htbp]
{\small
  \centering
  \caption{Impact of the numbers of control points and meta-handles. The last row is the model used in the rest of the experiments. `CD' indicates Chamfer distance, and `CotLap' indicates cotangent Laplacian.}
    \begin{tabularx}{\linewidth}{c|c|C|C}
    \toprule
    \footnotesize{\# Control Points} & \footnotesize{\# Meta-Handles} & CD $\downarrow$   & CotLap $\downarrow$ \\
    \midrule
    25    & 15    & 0.0644 & 0.5955 \\ \hline
    100   & 15    & 0.0715 & 0.9101 \\ \hline
    200   & 15    & 0.0758 & 0.8777 \\ \hline
    50    & 30    & 0.0621 & 0.8948 \\ \hline
    50    & 15    & 0.0628 & 0.5751 \\ 
    \bottomrule
    \end{tabularx}%
  \label{tab:control_point}%
}
\end{table}%

\subsection{Comparison with PointNet-based 3D Discriminator}
\label{sec:3d_discriminator}

\begin{figure*}[t]
\begin{center}
   \includegraphics[width=\linewidth]{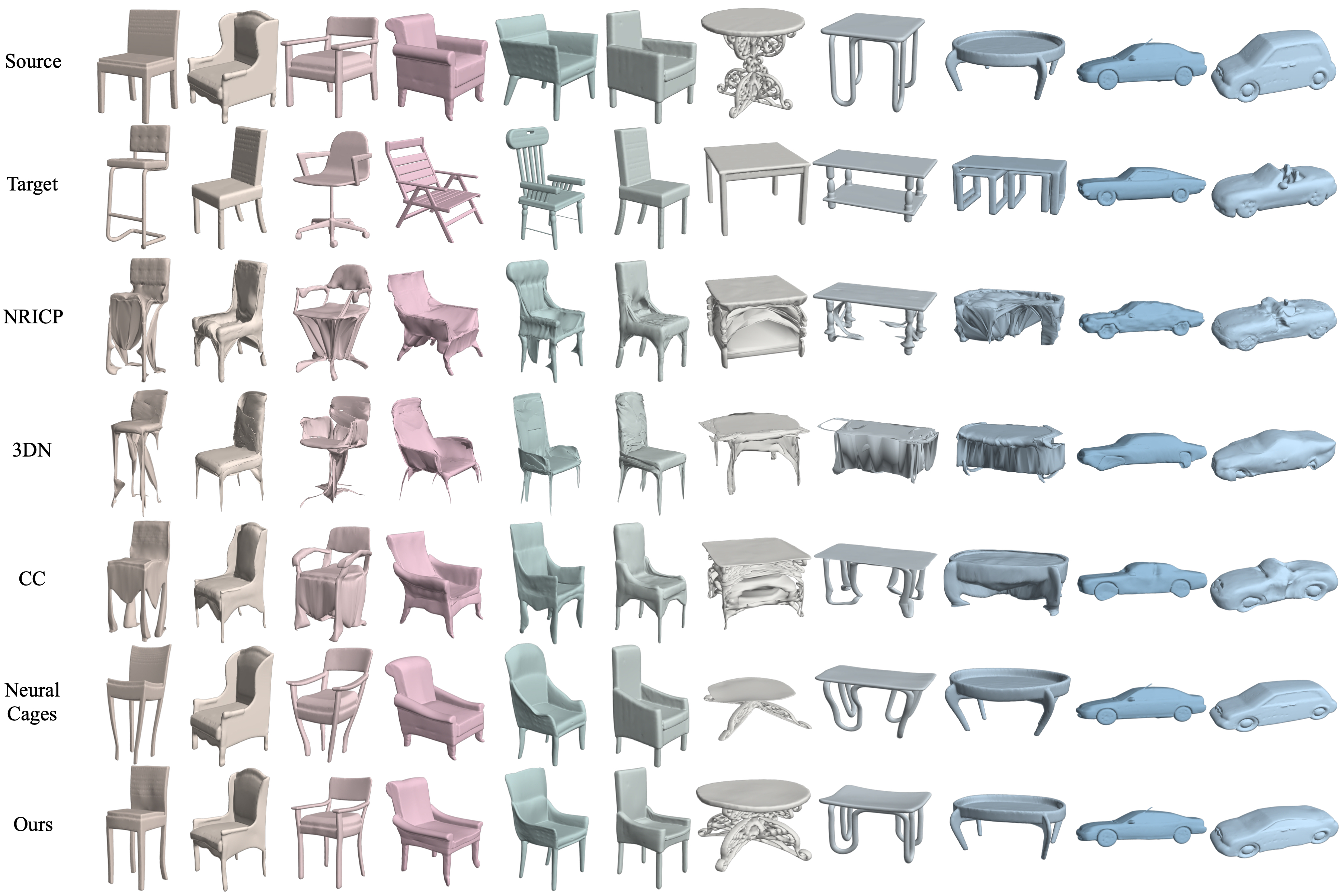}
   \caption{Qualitative comparison of our method with other deformation methods~\cite{huang2017learning,3DN,groueix2019unsupervised,NeuralCages}. (More examples of Figure~\ref{fig:pair-qualitative}~\refpaper~.)}
   \label{fig:sup_target_driven}
\end{center}
\end{figure*}

While we leverage a differentiable renderer and a 2D network for the discriminator, one can consider directly feeding the 3D deformed shape to a 3D processing network. To compare our discriminator with the case of directly processing 3D, we implemented another discriminator using PointNet~\cite{PointNet} and fed the points sampled over the deformed 3D mesh as input. Figure~\ref{fig:3dgan} shows some comparisons. Although the PointNet-based 3D discriminator can also prevent large distortions, we found that our 2D discriminator produces more visually pleasing deformations in practice. This might happen since the 2D discriminator can capture more subtle visual differences in the 2D space comparing with taking only account with the 3D geometry.

\subsection{Application: Data Augmentation}
\label{sec:supp_data_aug}

\begin{table}[htbp]
{\small
  \centering
  \caption{Data augmentation for subcategory classification.}
    \begin{tabularx}{\linewidth}{C|C}
    \toprule
          & Test Accuracy \\
    \midrule
    No Augmenation & 88.3\% \\ \hline
    w/o $\mathcal{L}_{adv}$ & 89.8\% \\ \hline
    Target-Driven & 90.4\% \\ \hline
    Ours  & 91.6\% \\
    \bottomrule
    \end{tabularx}%
  \label{tab:data_augmentation}%
}
\end{table}%

Our method learns a plausible deformation space for the input shape and can thus be used for data augmentation. We evaluate our approach as a tool of data augmentation with a multi-label shape classification task. Specifically, we use ten subcategories of ShapeNet~\cite{ShapeNet} chair models, and each model can belong to multiple subcategories (e.g., armchairs and swivel chairs). We sample 50 ShapeNet chair models as training data and 500 chair models as test data. To balance the data, while sampling, we ensure that each subcategory appears at least five times in the training data and at least 50 times in the test data. We employ PointNet~\cite{PointNet} as the classification network and train it with binary cross-entropy loss for each subcategory. We test four different settings: a) training on $50$ shapes without data augmentation; b) training on $50\times 20$ augmented shapes, where we utilize our method to randomly generate 20 variants within the deformation space of each shape; c) same with b) but without the adversarial loss $\mathcal{L}_{adv}$ when training our network; and d) training on $50\times 20$ augmented shapes, where we randomly sample 20 targets for each shape and use our target-driven deformation to generate the variants. For all the generated deformations, we keep their original subcategory labels for training. The results are shown in Table~\ref{tab:data_augmentation}. The results verify that our method can improve the classification accuracy and also that the adversarial loss $\mathcal{L}_{adv}$ is essential to generate plausible deformations. The target-driven deformation is less effective in sampling all the plausible variants.

\begin{figure}[t]
\begin{center}
   \includegraphics[width=0.8\linewidth]{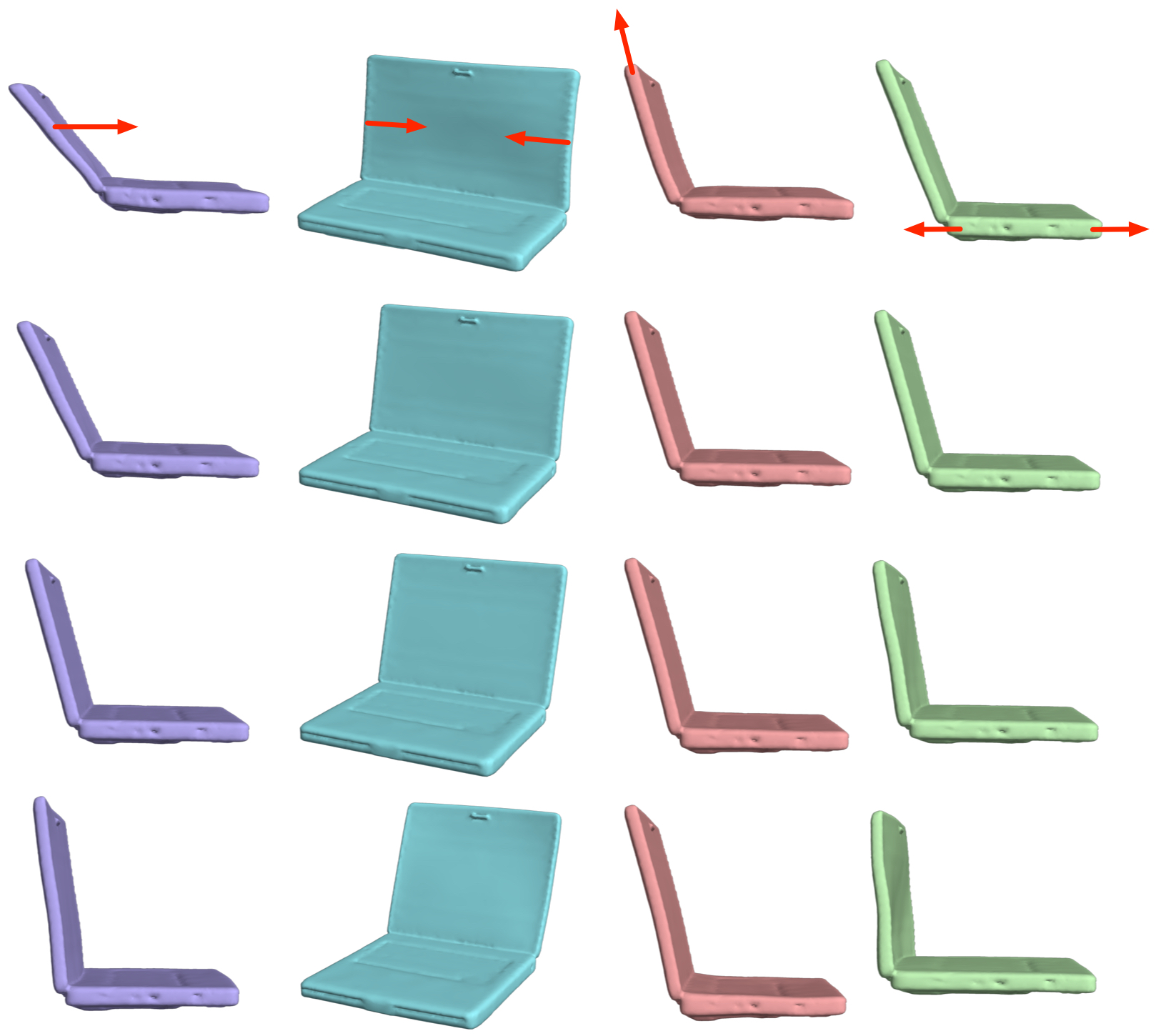}
   \caption{The learned meta-handles for the laptop category. Each column shows the deformations along the direction of a meta-handle.}
   \label{fig:laptop}
\end{center}
\end{figure}

\subsection{More Results of Target-Driven Deformation and Learned Meta-Handles on Laptops}
\label{sec:supp_more_target_driven}

Figure~\ref{fig:sup_target_driven} shows more results of the target-driven deformation, as also shown in Figure~\ref{fig:pair-qualitative}~\refpaper~.
Figure~\ref{fig:laptop} also shows the learned meta-handles on the laptop category. The results show that our method can learn the articulated motion of the two parts.